%% file: lineflow.tex
\documentclass{article}

\usepackage{microtype}
\usepackage{graphicx}
\usepackage{longtable}
\usepackage{subfig}
\usepackage{multirow}
\usepackage{multicol}
\usepackage{booktabs} 
\usepackage{enumitem}
\usepackage[ruled,vlined]{algorithm2e}
\usepackage{listings}

\usepackage{hyperref}

\usepackage[accepted]{icml2025}

\usepackage{amsmath}
\usepackage{amssymb}
\usepackage{mathtools}
\usepackage{amsthm}

\usepackage[capitalize,noabbrev]{cleveref}

\theoremstyle{plain}

\theoremstyle{definition}

\theoremstyle{remark}

\usepackage[textsize=tiny]{todonotes}

\icmltitlerunning{LineFlow: A Framework to Learn Active Control of Production Lines}

\newcommand{\RR}{\mathbb{R}}
\newcommand{\NN}{\mathbb{N}}
\newcommand{\Exp}{\mathrm{Exp}}
\newcommand{\cT}{\mathcal{T}}
\newcommand{\cS}{\mathcal{S}}
\newcommand{\cA}{\mathcal{A}}
\newcommand{\cF}{\mathcal{F}}
\newcommand{\Tsim}{T_{\mathrm{sim}}}
\newcommand{\Tac}{T_{\mathrm{AC}}}
\newcommand{\Tstep}{T_{\mathrm{step}}}

\definecolor{codegray}{gray}{0.95}
\definecolor{commentgray}{gray}{0.3}
\definecolor{keywordblue}{rgb}{0.26, 0.35, 0.76}
\definecolor{stringred}{rgb}{0.7, 0.1, 0.1}

\lstdefinestyle{mintedstyle}{
    backgroundcolor=\color{codegray},
    commentstyle=\color{commentgray}\itshape,
    keywordstyle=\color{keywordblue}\bfseries,
    stringstyle=\color{stringred},
    numberstyle=\tiny\color{gray},
    basicstyle=\ttfamily\small,
    breaklines=true,
    breakatwhitespace=true,
    showstringspaces=false,
    tabsize=2,
    numbers=left,
    numbersep=8pt,
    frame=single,
    framerule=0pt,
    rulecolor=\color{gray},
    xleftmargin=1.5em,
    xrightmargin=0.5em,
    escapeinside={(*@}{@*)}
}

\begin{document}

\twocolumn[
\icmltitle{LineFlow: A Framework to Learn Active Control of Production Lines}



\icmlsetsymbol{equal}{*}

\begin{icmlauthorlist}
\icmlauthor{Kai Müller}{hke}
\icmlauthor{Martin Wenzel}{hke}
\icmlauthor{Tobias Windisch}{hke}
\end{icmlauthorlist}

\icmlaffiliation{hke}{University of Applied Sciences Kempten, Kempten, Germany}

\icmlcorrespondingauthor{Tobias Windisch}{tobias.windisch@hs-kempten.de}

\icmlkeywords{Machine learning for manufacturing, active production line control, reinforcement
learning, production line simulation}

\vskip 0.3in
]



\printAffiliationsAndNotice{}  

\begin{abstract}
    Many production lines require active control mechanisms, such as adaptive
    routing, worker reallocation, and rescheduling, to maintain optimal performance.
    However, designing these control systems is challenging for various reasons, and
    while reinforcement learning (RL) has shown promise in addressing these
    challenges, a standardized and general framework is still lacking. In this work,
    we introduce LineFlow, an extensible, open-source Python framework for
    simulating production lines of arbitrary complexity and training RL agents to
    control them. To demonstrate the capabilities and to validate the underlying
    theoretical assumptions of LineFlow, we formulate core subproblems of active line control
    in ways that facilitate mathematical analysis. For
    each problem, we provide optimal solutions for comparison. We benchmark
    state-of-the-art RL algorithms and show that the learned policies approach
    optimal performance in well-understood scenarios. However, for more complex,
    industrial-scale production lines, RL still faces significant challenges,
    highlighting the need for further research in areas such as reward shaping,
    curriculum learning, and hierarchical control.
\end{abstract}

\section{Introduction}

At its core, manufacturing is about transforming raw materials into finished goods, often on a large scale. 
In most production systems, the necessary process steps are carried out by
work stations which each component has to traverse sequentially. 
Typically, the work steps necessary have to be applied in a fixed order, creating an interdependence between
stations in the sense that performance issues of one station directly affect stations
down- and upstream. 

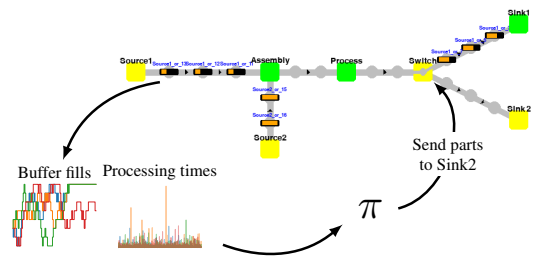
\begin{figure}[htbp]
    \centering
\scalebox{0.8}{
    \input{./images/overview.tex}
}
\caption{Active line control based on real-time data.}\label{f:active_control}\end{figure}

In theory, finding optimal layouts for production lines that take the interdependence of the individual work
steps into account is a computationally hard but well
understood problem called \emph{assembly line balancing
problem}~\cite{cost_oriented_line_balancing,state_of_the_art_simple_assembly_line_balancing}.
Solution strategies are powered by good stochastic models for interlinked production lines,
like in~\cite{bierboomsPerformanceAnalysisProduction2012,stochastic_modelling_of_manu_systems}, that
help computing the equilibrium state of the throughput for a given layout. 
However, in practice, optimal layouts and theoretic knowledge on the equilibrium of the throughput
does not guarantee that the production line unfold its full theoretical
potential. This may be due to situations like 
changing processing conditions, adverse coincidences in the
stochasticity of the processes, or machine failures. These circumstances can suddenly lead to
full buffers, jams, and shifting
bottlenecks~\cite{shifting_bottleneck,i40_and_bottleneck}. 
Thus, reacting in a dynamic and optimal way \emph{at runtime} is key to keep the performance of the
production line at its best.
At present, this is either a completely manual task accomplished by line workers on-site or by
\emph{line control} systems, that actively adjust line parameters in real-time (see
Figure~\ref{f:active_control}).
Possible interventions are rerouting tasks to alternative stations~\cite{Job_scheduling_based_on_BN_pred},
reallocating workers~\cite{worker_reallocation,worker_assignments}, or scheduling maintenance~\cite{rl_for_maintainance_planning}. 
Classical active line control systems typically combine rule-based strategies and mathematical models to
optimize scheduling and resource allocation, focusing mostly on bottleneck identification, see for
instance \cite{bottleneck_walk,active_period_method,Bottleneck_control,dynamic_bottleneck_detection} and references therein.
Finding optimal control strategies is a complex task that 
not only requires a detailed understanding of the production line dynamics but also involves time-intensive
trial-and-error session on the real system and simulations.
While effective in predictable scenarios, classical approaches often struggle in dynamic and
complex environments with high uncertainties, which has
motivated the use of machine learning for tasks like predicting bottlenecks~\cite{Review_ai_for_BN_pred} and
forecasting demand~\cite{demand_forecasts}.  

Since active production line control is, at its core, a sequence of observing a
state and taking action, it seems like a natural next step to learn line control
in an end-to-end fashion through interactions with a simulation using
reinforcement learning (RL). RL algorithms have demonstrated the ability to
learn complex relationships between states and
actions~\cite{alphastar,alphago,daydreamer,alphafold}. Significant
research has been conducted on learning control policies for production lines
(see Section~\ref{s:related_work}); however, progress is hindered because
existing studies each rely on their own ad hoc or domain-specific simulations,
making benchmarking, reproducibility, and the assessment of RL progress in
active line control difficult. One reason for this is the absence of a
theoretical or practical framework to train and evaluate RL agents in various
production settings in a standardized way.

In this work, we address exactly this research gap by
introducing LineFlow\footnote{\url{https://github.com/hs-kempten/lineflow}.}, a free and extensible
Python package specifically tailored to simulate production lines in order to
train RL agents for active line control at scale (Section~\ref{s:lineflow}). 
LineFlow not only is a fine-graded and fast discrete-event simulation of production lines,
it also comes with
an implementation of conventional action- and state-spaces as well as a flexible score
(Section~\ref{s:performance}) inspired by classical performance indicators that allows to train
agents for various production settings (Section~\ref{s:rl-formulation}).
To evaluate its theoretic underpinnings, we study well understood and realistic production scenarios by computing their
mathematical optimum (Section~\ref{s:case-studies}) and then
comparing it with the performance of control policies interacting with LineFlow
(Section~\ref{s:benchmarks}). To showcase the current limits, we also
analyze a very complex production setting as mentioned above with unclear optimum and train RL
agents via curriculum learning (Section~\ref{s:non-stationary}).
As there exist strong privacy concerns among manufacturers when it comes to releasing production
line layouts, publicly available datasets holding performance information and non-trivial layouts
are rare. Here, we hope that LineFlow will close this gap by allowing manufacturers to provide
synthetic non-confidential digital twins preserving their key runtime challenges to the public.
In Section~\ref{appendix:validation}, we model a real-world production line based on publicly
available data in LineFlow. Our research is purely application-driven by the needs of end-users in
manufacturing and opens up a wide range of challenges for machine learning researchers to
investigate. Since implementing active control systems is currently a highly manual task guided by
domain expertise, we aim for LineFlow to accelerate research on learning active
line control in the future.

\subsection{Related Work}\label{s:related_work}

There has been a lot of research where RL methods are studied to control a given
manufacturing technology, like for welding~\cite{rl_welding, rl_laser_welding_real_system,
rl_laser_welding_calibration} or molding~\cite{rl_molding_1,rl_molding_2} just to name a few. We refer to~\cite{rl_for_process_control} and
references therein for an
overview. In contrast, our work treats the industrial technology happening inside 
the processes as a black box and solely considers the \emph{processing time} of the stations
and their statistical interplay. This means, we neither model physical relations nor monitor
physical parameters of the processes, which is a fruitful application field for RL on its own~\cite{statistical_process_control_with_rl}.

A large body of research directly related to our work has been done for production scenarios where tasks of varying
length have to be scheduled on heterogeneous machines 
with different processing capabilities or resources. 
Such \emph{scheduling problems} are tackled with RL-based approaches in many works,
like in~\cite{scheduling_with_rl_1,scheduling_with_rl_2,rl_for_machine_scheduling,
active_control_with_rl,task_assignments_with_rl,scheduling_with_rl_semi_automated}. 
A comprehensive overview is provided in~\cite{scheduling_with_rl_overview}.
Other research focuses on machine interactions, as in~\cite{rl_2m1b}, 
where RL is used to control a layout consisting of two stations and one buffer to maximize
throughput or in~\cite{rl_for_energy_costs_1,rl_for_energy_costs_2}, where
energy costs are minimized by switching stations dynamically to standby mode.
In~\cite{rl_for_maintainance_planning},
the scheduling of predictive maintenance of a production line using RL has been studied.

\section{Simulating Production Lines}\label{s:lineflow}

Production lines can be characterized in many ways, like by the types of their processes or parts,
their production volume, or how their stations are
arranged~\cite{MachineLearningInProductionLiteratureReview2020,stochastic_modelling_of_manu_systems}. 
Here, we focus on production lines that produce discrete items. This section
introduces general principles and mechanisms of such production systems and explains how these are reflected in LineFlow.

\subsection{Objects of Production Lines}

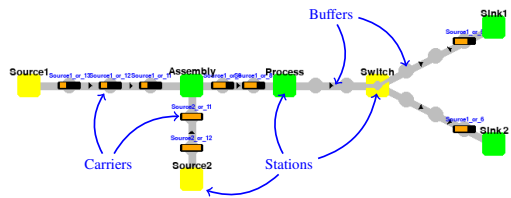
\begin{figure}
\centering
\scalebox{0.8}{
\input{./images/showcase.tex}
}
\caption{A production line visualized with LineFlow.}
\label{fig:showcase}
\end{figure}

The main objective of production lines studied in this work is to produce discrete items called \emph{parts}. 
Typically, the production of a single part requires a specific set of work steps that need to be applied in a
certain order involving the production and assembly of a series of sub-\emph{components}. Each work
step is carried out in a specific \emph{station}. 
One of the objectives of LineFlow is to model the statistical dynamics of a production line by considering the
individual processing times of the stations and their interplay.
To be more precise, the processing times are assumed to be exponentially distributed like
in~\cite{bierboomsPerformanceAnalysisProduction2012} as $\cT=T+\Exp_{S}$, 
where $T\ge 0$ is the
minimal processing time possible and where
$\Exp_{S}$ denotes
the exponential distribution\footnote{In the
literature typically denoted as $\Exp_{\frac{1}{\lambda}}$.} with mean $S$. 
We distinguish different types of stations, among them are 
\emph{sources} that set up components,
\emph{processes} that apply a piece of work on a single component, 
\emph{assemblies} that join two or more components into one, 
and \emph{sinks} that remove a fully
build part from the line (see Figure~\ref{fig:showcase}). 
Components are transported by \emph{carriers} from station to station via
\emph{buffers}, which work on a first-in-first-out principle. Buffers can only hold a predefined
number of carriers, called their \emph{capacity}. 
Time-intense work steps may be distributed over identical parallel stations,
while \emph{switches} handle the routing of carriers. The arrangement of stations and buffers is
typically called the \emph{layout} of the production line.

\subsection{Active Line Control}\label{s:activecontrol}

The overall performance of a production line depends sensitively on the individual performances of
its stations as performance issues of any station are propagated down- and upstream by
their linking buffers. For instance, consider the production line shown in
Figure~\ref{fig:showcase}: If the processing time of \emph{Process} is higher than the processing
time of \emph{Assembly}, the buffer in between fills up and once its maximal capacity is reached,
the
previous station is blocked as it cannot push finished components to the buffer. Such a jam can be
propagated backwards through the line and may even block the source. Similarly, subsequent stations like the two
sinks with faster processing times are not served with components and have to wait till new
components arrive. In the equilibrium, the performance of the full system depends essentially on the bottleneck
station, which is in simple cases just the station having the highest processing time. 
Buffers constitute an important parameter of production lines as they can dampen fluctuations in
processing times at the stations. However, buffers and their capacity are fixed parameters of the
layout, whereas stochasticity in the processing times of the processes can result in a shifting
runtime
bottleneck~\cite{shifting_bottleneck}. Thus, the bottleneck can vary, whereas
the buffer capacities cannot.
We refer to~\cite{bierboomsPerformanceAnalysisProduction2012} for a detailed analysis of the interplay
of stations and buffers. 
Bottlenecks in production lines do not only reduce throughput by slowing down the overall part flow,
but they can also lead to quality issues and even scrap. One prominent example arises in processes
with strict timing constraints between consecutive steps. In adhesive bonding, for instance,
components must be joined within a certain time window after the adhesive is applied. If a
downstream bottleneck causes delays and this threshold is exceeded, the bond can no longer be
guaranteed to meet quality standards, and the affected part must be
discarded~\cite{adhersive_bonding}. Such timing-sensitive constraints make bottleneck prevention
critical not only for efficiency, but also for product quality and material waste reduction.

To circumvent or mitigate such harmful runtime effects, we essentially study three
corrective actions and their interplay:
Changing worker assignments to speed up
bottleneck stations, changing the distribution of components, and including waiting times to prevent
scrap.  
On the one hand, increasing the waiting time is an important counter activity to prevent scrap, but there is a thin
line between preventing scrap and slowing down the bottleneck and, in turn, the whole line
(see Section~\ref{s:waiting-time}). 
Switches, on the other hand, allow to change the routing and distribution of components to keep
the bottleneck loaded constantly. For instance, if the processing time of \emph{Sink1} starts to
increase, the switch could change the distribution ratio between the sinks (see
Section~\ref{s:part-distribution}). 
Finally, reassigning
workers from one station to another directly impacts their processing time and can help to handle
peak loads but can create a bottleneck elsewhere (see Section~\ref{s:worker-assignments}). 
A more involved scenario where all these actions have to be combined effectively is studied in
Section~\ref{s:non-stationary}. 
Even if a bottleneck is correctly detected, choosing an optimal action sequence to mitigate is challenging
since actions take time to show results. 
For example, reassigning a worker requires task completion and relocation first.
While this reduces the bottleneck's mean processing time, high variance makes changes hard to
detect. In general, even with the right actions, high data variability delays recognizing
improvements.
Compounding this issue, production lines often operate near equilibrium states, where minor
disruptions can cascade into buffer overflows, bottlenecks, or system-wide jams. 
Moreover, the combination of different actions may have unintended
negative consequences on the performance. For instance, if 
\emph{Sink1} is the current bottleneck, then assigning more workers to \emph{Sink1} and sending more
components to \emph{Sink2} additionally, may result in a worse overall performance. 
To apply optimal actions, line control systems rely on real-time data from various production line
objects, which is often noisy, incomplete, and lagged (see Section~\ref{s:rl-data} and
Section~\ref{appendix:states}).

\subsection{Performance Measurements}\label{s:performance}

There are several ways to measure the performance of production lines.
A commonly used metric is the \emph{Overall Equipment Effectiveness}, short OEE, which quantifies the
effectiveness of equipment and machinery used~\cite{nakajima1988introduction}.
Assume we have given a control policy $\pi$ for a production line decomposed into stations $P_1,\ldots, P_k$.
Denote by $n^\pi_{\text{ok}}(t, i)$ 
and $n^\pi_{\text{nok}}(t, i)$ the cumulated number of OK and not OK (NOK) parts produced at station $P_i$ at
time $t$. 
Then,
following~\cite{nakajima1988introduction}, the OEE for $P_i$ is essentially defined as
$\text{OEE}_i(t) =
\frac{T_i}{t}\cdot n^\pi_{\mathrm{ok}}(t,i)$
where $T_i$ is the minimum of the processing time of $P_i$. By construction, we have
$n^\pi_{\mathrm{ok}}(t,i)\cdot T_i\le t$ for all $t$ and thus the maximal possible OEE is $1$.
Analogously, the OEE of a production line can be defined. The OEE definition is mainly focusing on
productivity and quality, but a more economic view requires to also take costs of the individual
stations and components into account, for instance as some stations require more energy for their
processing, like in~\cite{rl_for_energy_costs_1}, or because only some components are scraped, not
the full part. Thus, we assume that applying work at station $P_i$ comes with a cost $c_i$,
which may be material costs for joining a component or costs for energy as
in~\cite{rl_for_energy_costs_1}. The value of the final
part is considered to be the sum of all costs $c=\sum_{i=1}^kc_i$.
At simulation time $t$, the production has produced an aggregated value of
\begin{equation}\label{equ:costs}
C_\pi(t)=\frac{T_C}{t}\left(c\cdot n^\pi_{\text{ok}}(t) - \sum_{i=1}^kc_i\cdot
n^\pi_{\text{nok}}(t,i)\right)
\end{equation}
where $n^\pi_{\text{ok}}(t)\in\NN$ denotes the number of parts
produced at all sinks combined and $T_C$ denotes the minimal possible time the
line needs to produce a single part. One can think of Equation~\ref{equ:costs} as the value the
production line generates: If $C_\pi(t) > 0$, the line generates revenue, if $C_\pi(t)< 0$, it
generates costs. If no scrap can occur, maximizing $C_\pi(t)$ is equivalent to
maximize the number of produced parts. 

Generally, one wants to find a control policy $\pi$ that maximizes $C_\pi(t)$ and
\emph{simultaneously} minimizes the time $t$ to reach the maximum. It then depends on the exact usecase to weight a
smaller $t$ needed with a higher $C(t)$ reached by. To circumvent this distinction in our work and to create a
general benchmarking environment, one either has to fix the available time $\Tsim$ and to maximize the
output $C_\pi(\Tsim)$, or to fix the desired output $\tilde C$ and to minimize the time needed by
a policy $\pi$ to reach it. In our benchmark studies, we decided to fix $\Tsim$ as this typically
originates from high-volume production where the output needs to be maximized in a given work shift
of length $\Tsim$.

\section{RL for Active Line Control}\label{s:rl-formulation}

In this section, we describe how RL can be used to learn policies for active
line control. Specifically, we consider active line control as an \emph{episodic} and
\emph{partially-observed} Markov decision process. Let $\cS$ be a set of states and $\cA$ a set of
actions. Transition probabilities $P$ dictate how
likely
transition from a state $s\in\cS$ to a state $s'\in\cS$ is when choosing action $a\in\cA$ at $s$, namely
$P(s'|s,a)$. When selecting $a$ at $s$, a reward $R(s, a)$ is observed. 
Additionally given a desired length $T\in\NN$, this gives a \emph{Markov decision
process} $(\cS, \cA, P, R, T)$ where the goal is to find a \emph{policy} $\pi:\cS\to\cA$ to maximize
$\mathbb{E}_{P, \pi}\left[\sum_{t=0}^T R(s_t, \pi(s_{t-1})\right]$
where $(s_t)_{t\in\NN}$ is chosen under the interplay of $P$ and $\pi$.
If the states  $s\in\cS$ cannot be observed
directly by the policy but only features $f\in\cF$ instead controlled by another conditional
probability density function $O$, i.e., the likelihood of observing $f$ at state $s$ is $O(f|s)$,
then the tuple $(\cS, \cA, P, R, \cF, O)$ defines a \emph{partially-observed} Markov
decision process where a policy $\pi:\cF\to\cA$ needs to be found maximizing
$\mathbb{E}_{P,O, \pi}\left[\sum_{t=0}^T R(s_t, \pi(f_{t-1})\right]$.
Note that the definitions given here are streamlined versions of the common way of defining Markov
decision processes, where the process typically ends in a set of defined terminal states and
additionally, a \emph{discount factor} $\gamma\in[0,1]$ is given to trade-off rewards in early and
late states.

While the formulation above assumes a discrete-time decision process, we emphasize that the
underlying dynamics of the production line still can be \emph{continuous time} and only the agent
interaction—i.e., observation and action execution—is restricted to fixed intervals. This modeling
choice reflects real-world production control, where decisions like worker reallocation or routing
changes are typically made at regular intervals, such as every few seconds or at shift-level
granularity. This approach offers a practical compromise between physical realism and learning
efficiency, and is consistent with prior work in industrial reinforcement learning. 
Section~\ref{s:implementation} provides details how this is realized in LineFlow.

\subsection{Episodes and Rewards}\label{s:reward}

As described in Section~\ref{s:performance}, we consider constant operation times $\Tsim$ of the
manufacturing setting to be optimized. In principle, a control policy can interact with
a production line at any time.
We, however, assume that the policy can only interact
with the production line at fixed and equidistant
time points each $\Tstep$ apart, thus we get $T=\frac{\Tsim}{\Tstep}$ many interactions in an
episode. This approach introduces the possibility that the model may miss
certain state changes occurring between observation intervals, such as a very
brief buffer overflow.
However, from a RL point of view, this has multiple advantages: First, the trajectories all become
of fixed length and second,
the policy does not need the time scale. Allowing arbitrary interactions of a policy, particularly
at the early stage of the training, can lead to many thousands of interactions
within only a few time units. Although in LineFlow, $\Tstep$ and $\Tsim$ are considered to be
constants coming from the situation to be optimized, both can be varied sequentially in a curriculum
learning fashion~\cite{survey_cl}. The goal of an episode is to find a policy $\pi$ that maximizes $C_\pi(\Tsim)$.
To allow temporal
difference learning (see~\citep[Chapter~6]{rl_book}), we decompose $C_\pi$ into discrete quantities for $t\in\{0,\ldots, \Tsim\}$ by
defining 
$$R(s_t, \pi(s_{t-1}))=C_\pi(\Tstep\cdot (t+1))-C_\pi(\Tstep\cdot t).$$ 
That way, we get $\sum_{i=0}^T R(t)=C_\pi(\Tsim)$ as desired.

\subsection{States, Observations, and Actions}\label{s:rl-data}

Due to the stochastic nature of production processes and sensor inaccuracies, the true system state
of a production line is not directly observable. Instead, it must be inferred from incomplete,
noisy, and delayed observations. Key features relevant for performance estimation include buffer
fill levels, station processing times, production rates, station modes, and routing information from
switches~\cite{shifting_bottleneck,bottleneck_walk}. Whenever available, features in LineFlow are
constrained by known upper and lower bounds. For example, fill levels are normalized to $[0,1]$, while
production rates or processing times are positive but unbounded. 

The action space in LineFlow primarily consists of discrete control decisions, such as assigning
workers or routing components across stations. Additionally, stations can be switched on or off,
similar to~\cite{rl_for_energy_costs_2}. Workers are drawn from predefined pools associated with specific station groups. Each worker is
modeled as an individual and independent dimension in the action space, where the possible values
correspond to the stations the worker can be assigned to. While this introduces symmetries in the
action space when workers are indistinguishable, it enables fine-grained modeling of worker
attributes—such as skill levels or expertise—when they are not. Furthermore, representing workers as explicit
objects within the production line ensures that side constraints, such as a fixed number of
available workers, are respected at all times. Switches typically feature two discrete
actions: one controlling incoming buffers and another for outgoing buffers.
Beyond the core actions, LineFlow also supports modifications to the physical production
environment, such as actively adding or removing carriers from the line. While not included in our
case studies, these actions provide additional flexibility for modeling dynamic settings.
A list of all observable and actionable dimensions is in
Section~\ref{appendix:states}. 

\subsection{Implementation Details}\label{s:implementation}

LineFlow is designed to address key challenges in optimizing active line controls while meeting the
needs of RL researchers. The discrete event simulation, handling station interactions and
stochasticity, is built on SimPy~\cite{simpy}. Its object-oriented structure enables easy
customization of stations and rapid setup of complex production scenarios.
A visualization module based on pygame provides insights into layout dynamics and agent
interactions. Fully implemented in Python, LineFlow integrates seamlessly with data science tools
like pandas~\cite{pandas} and numpy~\cite{numpy}. RL interaction follows the gymnasium API~\cite{gymnasium}, allowing training
with stable-baselines3~\cite{stable-baselines} or skrl~\cite{skrl}. Environments support
vectorization and parallelization to accelerate training. Further details are
provided in Section~\ref{appendix:implementation}, with a complete example in
Section~\ref{appendix:example}, illustrating the layout in
Figure~\ref{fig:showcase}. Although agent interaction follows the discrete-time process described in
Section~\ref{s:rl-formulation}, the underlying production dynamics in LineFlow is
simulated in continuous time using a discrete-event engine. The interaction frequency of the agent
can be adjusted via a parameter, allowing the discrete-time control to approximate
continuous-time behavior with high fidelity.

\section{Case Studies}\label{s:case-studies}

In this section, we introduce and study realistic production scenarios where active control is required.
The scenarios have been selected for two reasons: First,
their theoretical optimum can be computed statistically allowing us to quantify whether RL
algorithms learn optimal policies. Second, they appear as subproblems in many production scenarios.

\begin{figure}
\centering
\subfloat[$\mathrm{WT}$ and $\mathrm{WTJ}$]{\label{f:waiting_time}
\includegraphics[width=0.15\textwidth]{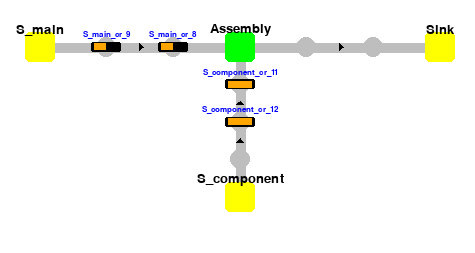}}
\hfill
\subfloat[$\mathrm{WA}_{3, 10}$]{\label{f:worker_assignment}
\includegraphics[width=0.09\textwidth]{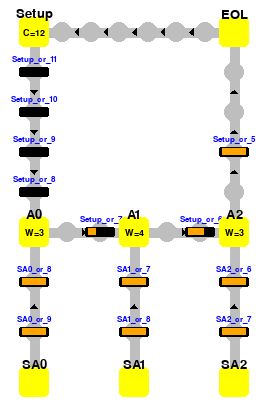}}
\hfill
\subfloat[$\mathrm{PD}_5$]{\label{f:multiprocess}
\includegraphics[width=0.15\textwidth]{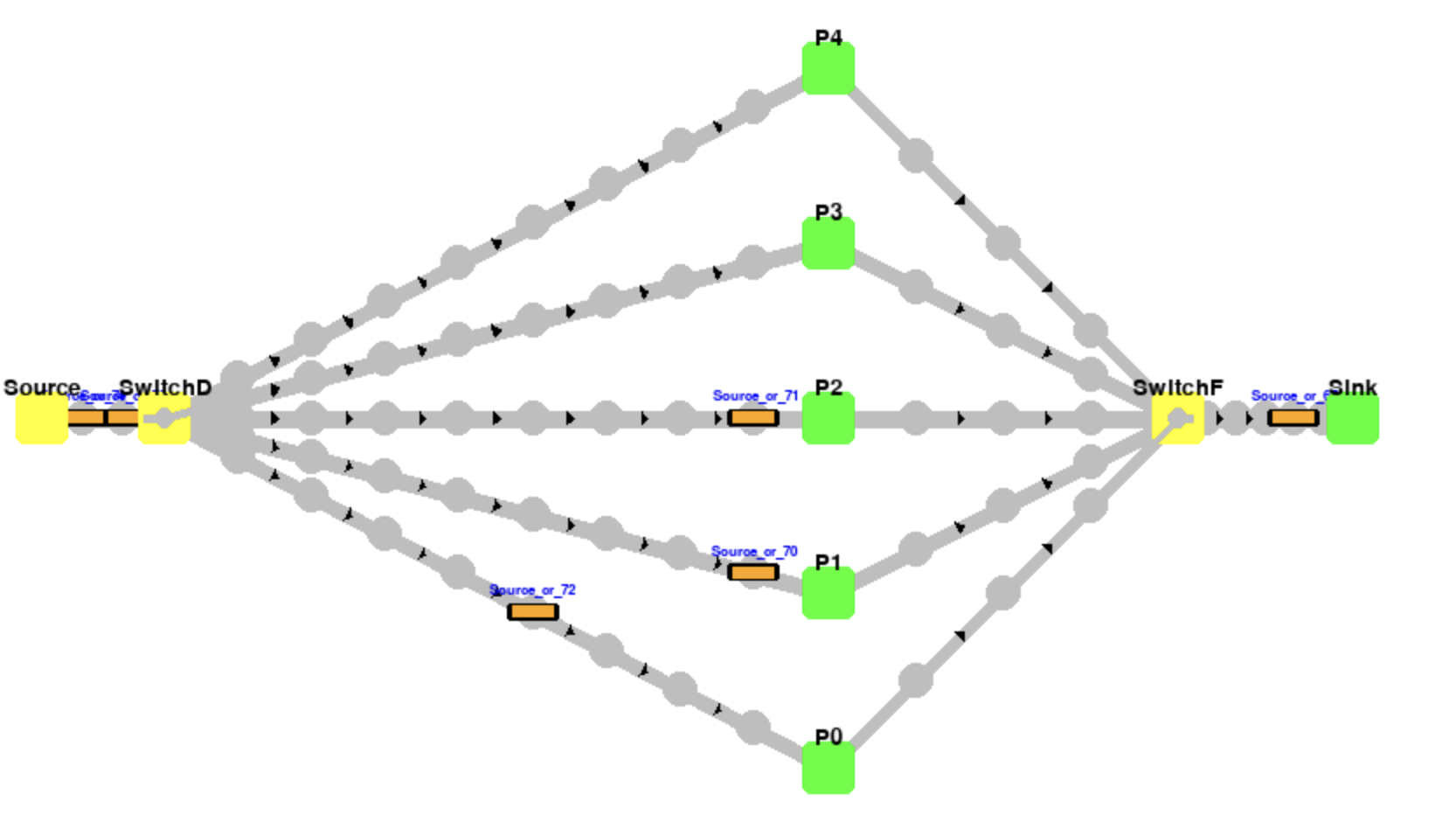}}
\caption{Representatives of the three atomic production line challenges analyzed in
this case study.}
\label{fig:case_studies}
\end{figure}

\subsection{Optimal Waiting Time $\mathrm{WT}$ and $\mathrm{WTJ}$}\label{s:waiting-time}

In this scenario, the optimal waiting time between parts produced by
a source station $S_C$ has to be found (see also Figure~\ref{f:waiting_time}). The source $S_C$ serves together with another source~$S_M$
an assembly station $A$, which
joins the components from both sources to produce the final product and sends it to sink~$S$.
Components from source $S_C$ have a fixed expiration time $\Tac$ called \emph{assembly condition}: 
If the time from their setup at $S_C$ to the time the process starts at~$A$
is larger than $\Tac$, then $A$ has to dispose this component and has to wait for the next one.
The removal of a scrap part not only generates costs, it
also blocks the
assembly for a certain amount of time as the defective component has to be removed first.
Such a situation is typical in many joining processes, like
gluing applications, where adhesives can dry out if not processed in time~\cite{adhersive_bonding}.
If the waiting time is chosen
too small, $S_C$ produces to many components which cannot be handled by $A$ in time and thus drives
scrap costs. On the other hand, if the waiting time is chosen to high, $A$ has to wait for components
which delays the production of final parts. The goal is to balance the waiting time of $S_C$ to maintain a
continuous component supply at~$A$. 
Assuming costs $c_M$ and
$c_S$ for components produced at $S_M$ and $S_C$
respectively
and that only components of $S_C$ can potentially be scraped, the performance of the line as
described in Section~\ref{s:performance} within a given time frame $t$ is
$$C_\pi(t)=\frac{T_C}{\Tsim}\left((c_M+c_S)\cdot n_{\mathrm{ok}}^\pi (t) - c_S\cdot n_{\mathrm{nok}}^\pi
    (t, A) \right).$$
We generally assume that the processing times without potential waiting times are such that
$A$ is the bottleneck of the line. 
Thus, the maximum number of parts produced depends on the time
$A$ needs to get one carriers and one component,
to assemble them, and to push the final product to the buffer. The optimal waiting time essentially
fills the between the times $A$ and $S_C$ need to handle and process their parts. We give an
explicit equation for the optimal waiting time and the maximum number of parts in Section~\ref{appendix:waiting-time}.

\begin{figure}
    \centering
    \includegraphics[width=0.45\textwidth]{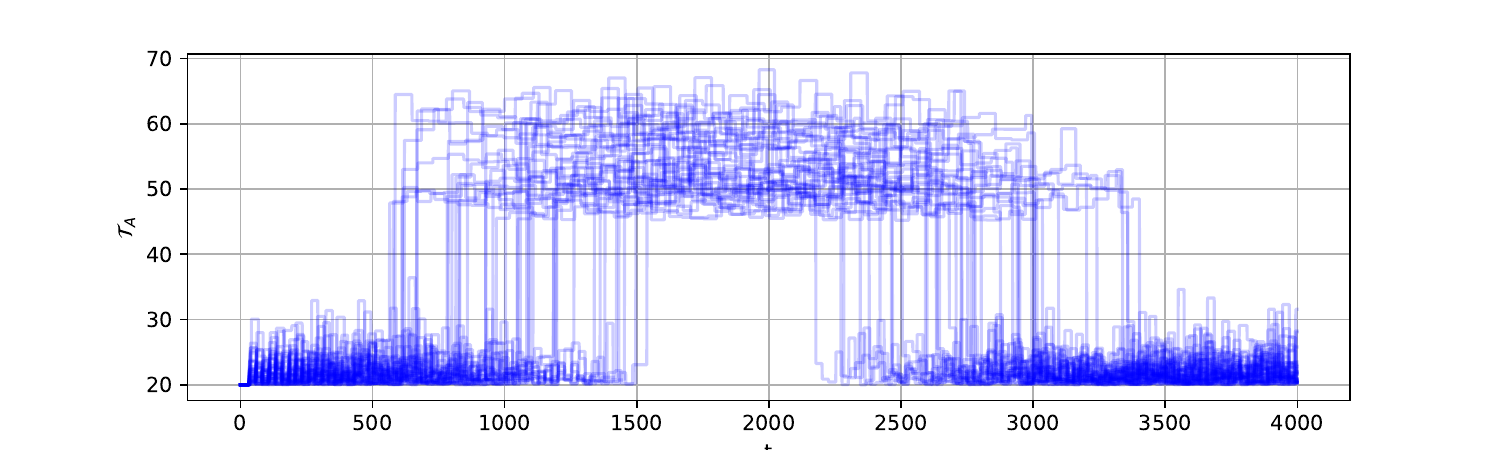}
    \caption{The jumps in the processing time of the assembly $A$ in $\mathrm{WTJ}$ for different simulations of
    length $4000$.}\label{fig:processing_time_jumps}
\end{figure}

To force a dynamic adaption we now introduce the related scenario
 \emph{Waiting time jump} ($\mathrm{WTJ}$).
Here, we 
uniformly sample a length $T_{\mathrm{jump}}$ and a trigger
time $T_{\mathrm{trigger}}$ such that $[T_{\mathrm{trigger}}, T_{\mathrm{trigger}} +
T_{\mathrm{jump}}]\subset [0, \Tsim]$. Then,
the processing time of $A$ at simulation time $t$ from
$$
\begin{cases}
    T+\Exp_S,&\quad\textnormal{ if }t\not\in[T_{\mathrm{trigger}}, T_{\mathrm{trigger}}+T_{\mathrm{jump}}]\\
    f\cdot T+\Exp_S,&\quad\textnormal{ if }t\in[T_{\mathrm{trigger}}, T_{\mathrm{trigger}}+T_{\mathrm{jump}}]
\end{cases}.
$$
As $f$ and $T_{\mathrm{jump}}$ can be different in each episode, the maximal possible reward can
vary, too. To ease the comparison of agents over multiple runs, we construct $f$ for given $T_\mathrm{jump}$ in a way that the
maximal possible reward
remains a constant (see Section~\ref{appendix:waiting-time} for details). More precisely, we fix a
constant $0.5<R<1.0$ and
construct $f$ such that expected maximal parts produced is $R\cdot N$, where $N$ is the expected
number of parts of $\mathrm{WT}$ without jump.
A visualization of the processing time of $A$ for multiple simulations is in
Figure~\ref{fig:processing_time_jumps}.
Although, by design, the expected maximal number of parts
produced by the line is $R\cdot N$, this value cannot be reached by a control agent. The reason is that
at time $T_\mathrm{trigger}$ if the processing time of the assembly jumps to a higher level, the new
processing time of $A$ can first be observed once the first part has been
produced. Thus, from $T_\mathrm{trigger}$ to $T_\mathrm{trigger}+f\cdot T+\Exp_S$, the source $S_C$
runs with a too low waiting time and possibly sends components to $A$ which are going to expire. The
same happens once the processing time of $A$ jumps to the lower level. Here, the source may run at
a higher waiting time causing $A$ to wait. 
To quantify the optimum reachable
by agents learning from observations, we evaluate the performance of an agent 
learning the means of the processing times online (see Section~\ref{appendix:waiting-time}).

\subsection{Optimal Part Distributions $\mathrm{PD}_k$}\label{s:part-distribution}

Our next case study is a subproblem of scheduling problems
in manufacturing and is already extensively researched in literature (see Section~\ref{s:related_work}).
Here, components have to be distributed onto $k$ many parallel processes $P_1,\ldots, P_k$ with processing time distributions
$\cT_1,\ldots,\cT_k$
by a single switch in an optimal way. 
Moreover, another switch needs to fetch the outputs of the processes and sends them to a sink (see
Figure~\ref{f:multiprocess}). We assume that $\cT_i=T_i+\Exp_{S_i\cdot T_i}$ where
$T_i$ and $S_i$ are constants for all $i\in[k]$. 
To compute the optimum, we assume for simplicity that get and put times from the connecting buffers are already
contained in $\cT_i$. Within a given time
frame $\Tsim$, the expected maximal number of parts $\mathcal{N}_i$ produced by process $i$
is obtained when $P_i$ is served with parts constantly without time
gaps:
$\mathbb{E}[\mathcal{N}_i]=\frac{\Tsim}{\mathbb{E}[\cT_i]}=\frac{\Tsim}{(1+S_i)\cdot T_i}$.
When we assume that the processing times of the source and the sink are negligible compared to the
processing times of the processes $P_i$. The maximal
number of parts $\mathcal{N}$ produced by the production line within
$\Tsim$ is the sum of the maximal parts
produced by all processes, i.e.,
$\mathcal{N}=\sum_{i=1}^k\mathcal{N}_i$ as all processes can work in
parallel. 
This allows us to compute the maximum number of parts
the layout can produce theoretically, see Section~\ref{appendix:optimality-part-distr}.
Clearly, the greedy policy deployed at both switches which pushes components on the buffer with lowest
fill and fetches components from the buffer with highest fill loads the processes in an optimal way
(see also Section~\ref{appendix:optimality-part-distr}). 

\subsection{Optimal Worker Assignments $\mathrm{WA}_{k,N}$}\label{s:worker-assignments}

In many production scenarios, processes involve manual work which can be completed faster by
multiple workers collaborating. In this case study, the goal is to find the right
distribution of a limited number of workers over multiple stations each having a varying processing
time.
We assume that the required processing time of a station, where $n$ workers are assigned to, is sampled from
the distribution
$\cT_{T, S, n}:=T\cdot p_c(n)+\Exp_{S\cdot T}$
where $T,S\in\RR_{>0}$ are constants and where
$p_c(n)=\exp(-c\cdot n)$
is a \emph{performance coefficient} (see also Section~\ref{appendix:optimality-worker}). 
In our benchmarks, we set $c=0.3$, which means that an additional worker reduces the
processing time by approximately $74\%$.
Now, consider a layout of $k$
sequential stations $A_1,\ldots, A_k$ with constants
$T_1,\ldots,T_k$ and $S_1,\ldots, S_k$. Moreover, assume the line has
a pool of $N$ workers to distribute. As the stations are sequentially arranged and connected via buffers (see
Figure~\ref{f:worker_assignment}), the slowest process determines the speed of the
overall production line. 
This means, a partition of $N$ into $k$ integer
summands, i.e., $n_1+n_2+\ldots + n_k=N$ with $n_i\in\NN_0$ needs to
be found such that the maximum of all processing times 
is minimal. Clearly, the assignment process is bound to certain
constraints, like delayed start of
processes due to the waiting for assigned workers because of traversal
times. We refer to Section~\ref{s:rl-data} for more
details about how the worker assignment is modeled in LineFlow.
Let $\mathcal{N}_{N,k}\subseteq[N]^k$ be the set of
partitions of $N$ into many $k$ non-negative integer numbers. More formally, the following max-min
integer optimization problem needs to be solved:
\begin{equation}\label{equ:worker-assignment}
    T_C^* = \min_{(n_1,\ldots,n_k)\in\mathcal{N}_{N,k}} \max_{i\in[k]}\mathbb{E}[\mathcal{T}_{T_i, S_i, n_i}]
\end{equation}
where in our case, the expected value of the processing time is
$\mathbb{E}[\mathcal{T}_{T_i, S_i, n_i}]=T_i\cdot
p_c(n_i)+S_i\cdot T_i$. The maximum number of parts that can be produced is
then essentially given by $\frac{\Tsim-T_{\mathrm{off}}}{T_C^*}$ after removing some initial
offsets $T_{\mathrm{off}}$ due to the ramp-up of the production line.
Mathematically, solving Equation~\eqref{equ:worker-assignment} is computationally hard 
not alone because it is a min-max problem but also because it involves integer partitions. 
Learning an optimal
distribution in an interactive environment is even more challenging, as first good estimates for
$T_i$ and $S_i$ need to be learned as well as for $p_c$.
In Section~\ref{appendix:optimality-worker}, we formulate Equation~\ref{equ:worker-assignment} as
mixed-integer optimization problem and compare the empirical results obtained with LineFlow with the
theoretical optimum.

\subsection{Complex Line $\mathrm{CL}$}\label{s:non-stationary}

\begin{figure}
    \centering
    \includegraphics[width=0.45\textwidth]{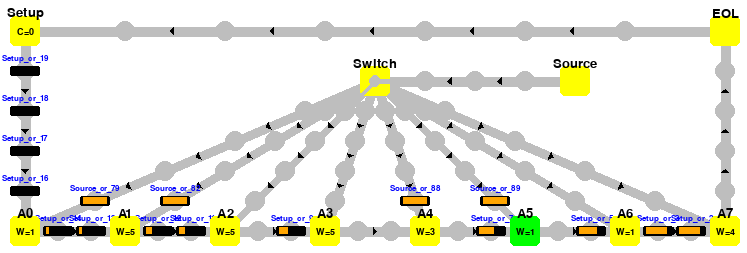}
    \caption{Scenario $\mathrm{CL}$ with $8$ assemblies.}
\label{f:complex}\end{figure}

Finally, we combined all studied problems into a problem we call
\emph{complex line}, short $\mathrm{CL}$ (see Figure~\ref{f:complex}). Here the agent has to distribute workers and
components over $k$ sequential assembly stations such that an assembly condition $\Tac$ as in
$\mathrm{WT}$ for the components is kept.
Interestingly, simply combining the individual and optimal solutions of the individual
subproblems does not lead to an optimal solution for~$\mathrm{CL}$. A straightforward distribution 
approach with a fixed distribution of workers, parts and a fixed waiting time
at the source can cause excessive scrap or a slow buffer-filling process between
the source, switch, and assembly stations.
In Section~\ref{appendix:complex-line}, we present a profitable heuristic found by extensive testing
meaning the number of produced parts outweighs the cost of scrap.

\section{Benchmarks}\label{s:benchmarks}

\subsection{Layout of Experiments}\label{s:experiments}

\begin{figure}
    \centering
\includegraphics[width=0.45\textwidth]{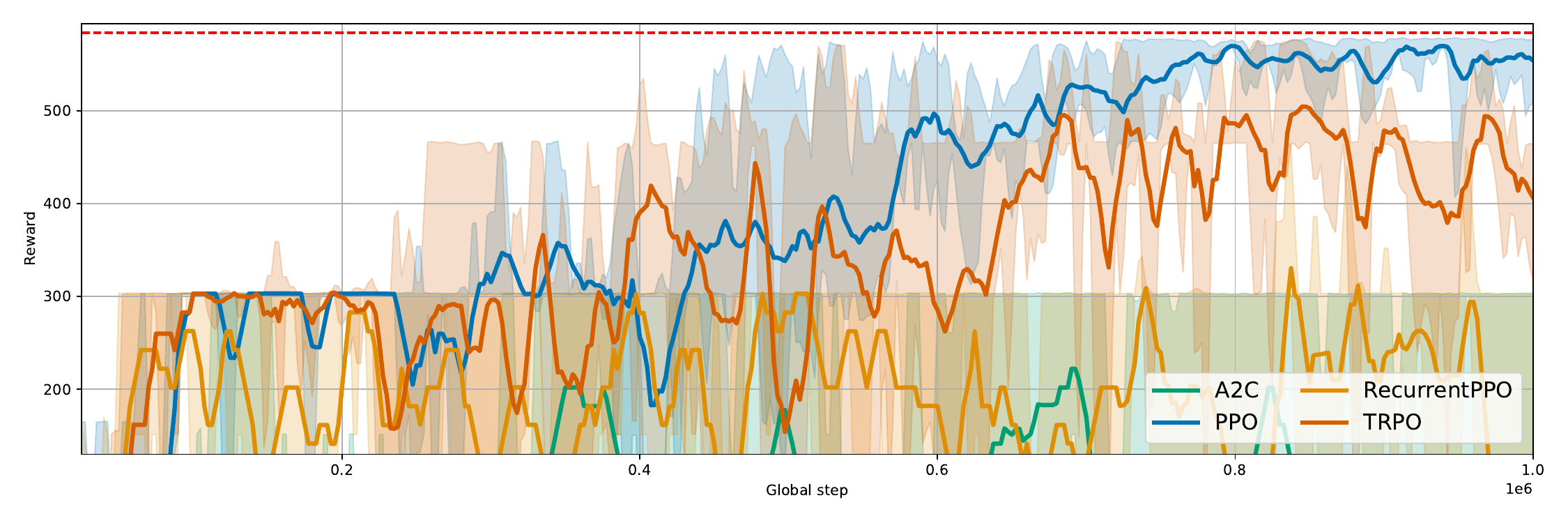}
\caption{Evaluation reward over global steps for $\mathrm{PD}_3$.}\label{f:reward_evolution}
\end{figure}

First, we want to emphasize that our benchmark is intended to serve as a proof of
principle how evaluation can be done with LineFlow and to show that optimal control can be learned.
For that, we only select a handful of algorithms and hyperparameters to consider, although many more
would be feasible as well.
Concretely, we use the following algorithms in our
experiments: PPO~\cite{rl_ppo} and its recurrent
version~\cite{rl_recurrent_ppo}, A2C~\cite{rl_a2c}, and
TRPO~\cite{rl_trpo}, all of which can deal with
multi-dimensional observations of mixed types and multi-dimensional
categorical action spaces. We used the implementations provided by the
\texttt{stable-baselines}~\cite{stable-baselines} package and
evaluated multiple hyperparameters with three different random seeds for each case study. More
details on the hyperparameters used can be found in Section~\ref{appendix:hyperparameters}.
The performance of the agents was measured in online evaluations, reporting the mean reward of $5$
episodes on a separate evaluation environment $5$ times (see also Figure~\ref{f:reward_evolution}
for the evolution of the reward over global steps).
When not stated differently, the
deterministic version of a trained policy is evaluated, that is, the action with the highest
probability is selected.
In all experiments, the agents are trained and evaluated on five environments stacked into a
vectorized environment.

\subsection{Results for $\mathrm{WT}$, $\mathrm{WTJ}$, $\mathrm{WA}$, and $\mathrm{PD}$}

All benchmarks we set $\Tsim=4000$.
For $\mathrm{WT}$ and $\mathrm{WTJ}$, we set $\cT_A=20+\Exp_{2}$,
$\cT_{S_C}=5+\Exp_{0.5}$, and use a constant get time $\cT_g=1$. The trigger time $T_\mathrm{trigger}$
is uniformly sampled from the interval $[500, 1500]$ whereas $T_\mathrm{jump}$ is uniformly sampled
from $[1600, 2000]$. The assembly condition is set to $\Tac=35$.
In $\mathrm{PD}_k$, the time distributions are $\cT_i=10\cdot(i+1)+\Exp_{0.1}$ for $i\in[k]$.
In $\mathrm{WA}_{k,N}$, we set $N=3\cdot k$ and use $c=0.3$ for the performance coefficient.
Moreover, we set $T_i=(16+i\cdot 4)$.

The performance of the best hyperparameter combination for each algorithm, averaged over multiple
seeds, is shown in Table~\ref{tab:Benchmarks}. More details of the rewards obtained during online evaluation of
all hyperparameter sets is in Section~\ref{appendix:hyperparameters}.
Scenario $\mathrm{WT}$ is feasible for all algorithms, but only recurrent PPO approaches the
optimal value for $\mathrm{WTJ}$. The actor-critic method A2C outperforms policy-based methods in
$\mathrm{WA}$ by achieving a
better performance with fewer steps. However, in part distribution, actor-critic methods fail to
learn good controls while policy-based approaches succeed.

\begin{table}
    \resizebox{0.48\textwidth}{!}{%
    \centering
    \begin{tabular}{l|llll|r}
        \toprule
         &       \textbf{PPO} & \textbf{TRPO} & \textbf{rPPO} & \textbf{A2C} & \textbf{Optimal}\\ \midrule
$\mathrm{WT}$        & $155.0\pm 3.0$ ($158.4$) & $157.1\pm 1.9$ ($158.6$) & $157.7\pm 1.3$ ($\mathbf{159.0}$) & $154.4\pm 5.9$ ($158.6$) & $156.2\pm 1.5$ \\ \midrule
$\mathrm{WTJ} $ & $93.7\pm 6.9$ $(101.4)$ & $93.9\pm 7.6$ ($102.4$) & $105.9\pm 11.6$
($\mathbf{113.0}$) & $100.5\pm 11.1$ ($110.2$) & $114.8\pm 0.9$ \\ \midrule
$\mathrm{WA}_{3,9}$  & $274.7\pm 13.3$ ($288.6$) & $244.6\pm 18.0$ ($262.8$) & $273.7\pm 21.9$ ($288.4$) & $278.9\pm 17.4$ ($\mathbf{289.4}$) & $287.1\pm 2.7$ \\ \midrule
$\mathrm{WA}_{4,12}$ & $243.1\pm 9.0$ ($252.0$) & $194.6\pm 7.7$ $(202.6$) & $231.9\pm 21.3$ ($246.4$) & $244.7\pm 15.4$ ($\mathbf{255.0}$) & $252.8\pm 2.1$ \\ \midrule
$\mathrm{WA}_{5,15}$ & $210.5\pm 22.0$ ($\mathbf{235.0}$) & $151.2\pm 6.9$ ($155.6$) & $211.9\pm 27.8$ ($230.0$) & $207.7\pm 38.4$ ($234.2$) & $236.3\pm 2.1$ \\ \midrule
$\mathrm{PD}_3$      & $578.7\pm 0.6$ ($\mathbf{579.2}$) & $539.1\pm 61.3$ ($577.8$) & $359.6\pm 95.1$ ($469.4$) & $304.3\pm 0.5$ ($304.8$) & $582.7\pm 2.6$ \\ \midrule
$\mathrm{PD}_4$      & $651.1 \pm 9.3$ ($\mathbf{661.2}$) & $582.1 \pm 101.5$ ($657.2$) & $585.8\pm 31.7$ ($620.2$) & $305.1\pm 0.1$ ($305.2$) & $670.7\pm 2.2$ \\ \midrule
$\mathrm{PD}_5$      & $651.3 \pm 107.0$ ($\mathbf{715.6}$) & $664 \pm 77.0$ ($712.6$) & $305.1\pm
0.2$ ($305.4$) & $358.3\pm 92.5$ ($465.2$) & $738.3\pm 3.5$ \\ \bottomrule
        \bottomrule
\end{tabular}}
\caption{Overview of rewards obtained by the best hyperparameter combination. Number in brackets denotes maximal
reward obtained (see Section~\ref{appendix:optimality} for details). 
Note that optimal policies are shaped by stochasticity as well and trained RL agents may
outperform the expected mean performance in individual episodes.}\label{tab:Benchmarks}
\end{table}

\subsection{Results for $\mathrm{CL}$}

\begin{figure}
    \centering
    \includegraphics[width=0.35\textwidth]{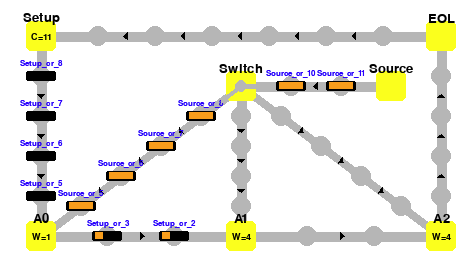}
    \caption{An agent trained with the PPO algorithm causing a blocking state of the line: All
    components send to $A_0$ causing $A_1$ to wait. As $A_1$ is
    blocked, the buffer between $A_0$ and $A_1$ gets full, which in turn blocks $A_0$ and
    consequently the full line.}
\label{fig:blocking_cl}
\end{figure}

\begin{figure}
    \includegraphics[width=0.5\textwidth]{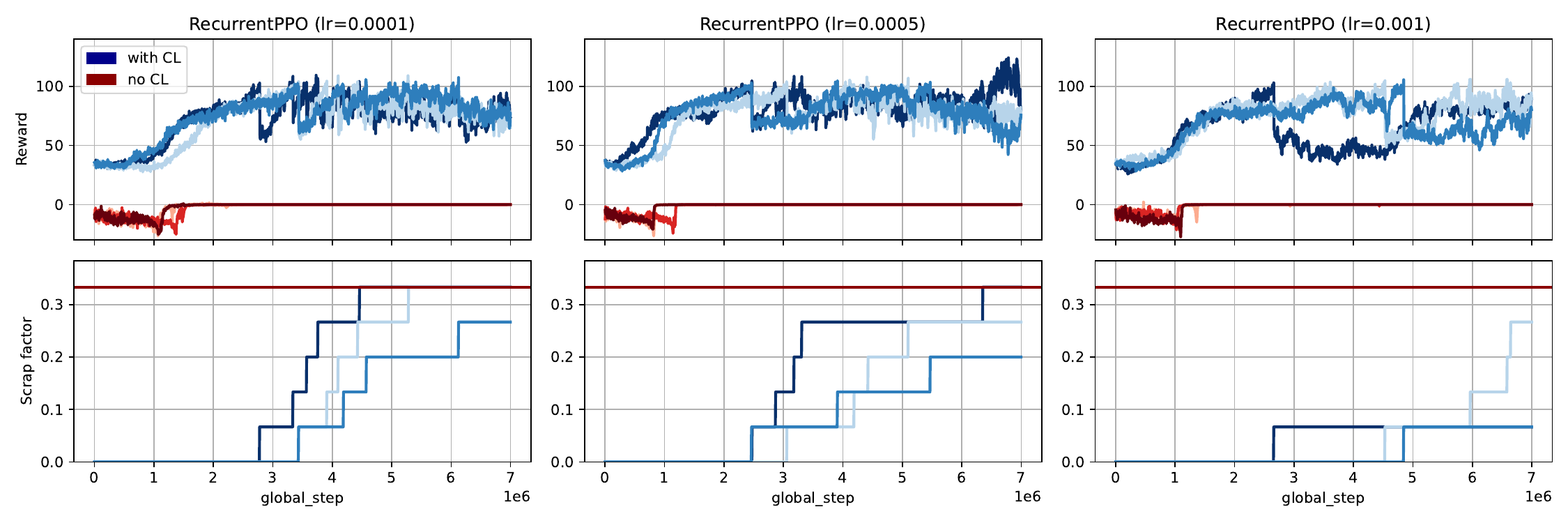}
    \includegraphics[width=0.5\textwidth]{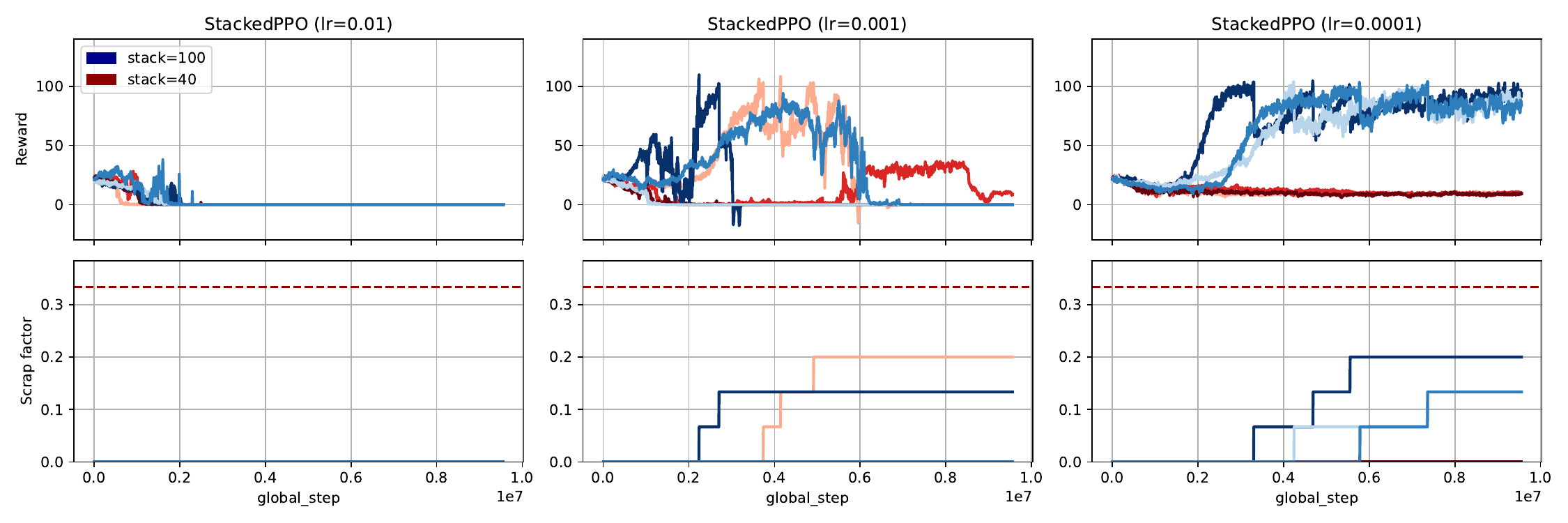}
    \caption{Upper figure shows a comparison of recurrent PPO trained with (blue) and without (red) curriculum
    using three different seeds. Lower figure shows a comparison of the stacked PPO trained
using curriculum with a stack of $40$ (red) and $100$ (blue).}
    \label{fig:cl_vs_no_cl}
\end{figure}

Informed decisions for $\mathrm{CL}$ require to keep past actions in memory. For instance,
sending a component to station $i$ at a give step
requires to send a part to station $i+1$ in the one of the subsequent step in order to keep the part flow on the main
track.
We found that, no matter which hyperparameters have been used, hardly an algorithm
reached a strictly positive reward. Since, at the start of the training the number of scrap parts heavily
outweighs the number of produced parts the reward starts significantly
negative. Interestingly the agents all converged to a reward of zero.
An investigation revealed that agents learn to dead-lock the line, leading to no
produced parts but also, no scrap parts (see
Figure~\ref{fig:blocking_cl}). From there, the same train data is generated during all subsequent
rollouts leading to poor update steps of the agent. In our experiments, no algorithm tested managed to escape from
these dead-lock situation when trained from scratch.
Thus, we applied a curriculum which slightly increases the scrap costs of a component scraped at one
of the assemblies. 
The intention behind is that first, the agents learn the general mechanics of the
production line and which actions are necessary to bring parts to the final station. Then, they have
to get more efficient by reducing the number of scrap parts. 
We increase the scrap weight by a constant factor until $\frac{1}{k}$ is reached
(in our experiments $0.006$ for
the scenario having $k=3$) once the reward is above $100$ in five subsequent evaluations.
As optimizing $\mathrm{CL}$ requires to memorize past actions, we consider only two algorithms here:
A stacked PPO~\cite{rl_ppo} receiving the past $k$ observations (with $k=40$ and $k=100$) and a
recurrent PPO~\cite{recurrent_policies} as in~\cite{memorygym}.
First, we can see that using an curriculum is \emph{effective}: 
Agents obtained through a curriculum obtain a large positive reward right from the
beginning and they manage to keep it also when the scrap costs
are increased. Figure~\ref{fig:cl_vs_no_cl} shows a comparison of
the recurrent PPO trained with and without
the curriculum for different random seeds and learning rates. In addition, we see that the stacked
version of PPO does not manage to reach the performance of the recurrent version, 
even when trained $10$ times longer.

When implementing a baseline for $\mathrm{CL}$, it is essential to consider the waiting
time of the source as well as the distribution of components and workers. Our heuristics prioritized the buffer with
the lowest fill level fills, while buffers feeding into later assembly stages receive higher
priority. We then conducted a grid search over various waiting times and worker distributions. The best
heuristic we identified achieved a reward of $254.6 \pm 1.1$. Overall, even the best-performing RL
algorithms in our $\mathrm{CL}$ benchmark still fall short of the manually implemented heuristic.

\section{Discussion}
In this work, we explore a rich problem class from manufacturing and
make it accessible for RL research by introducing a novel
framework LineFlow. 
We demonstrated that RL algorithms can learn effective control
policies for various production line challenges, achieving optimal
performance in well-understood scenarios. Moreover, we showed that
for more complex, dynamic production settings, traditional RL 
approaches struggle without additional techniques such as curriculum learning
and memory-based policies. We found that learning effective control 
strategies from interactions alone is
difficult as the ramp-up phase
of production plays a crucial role in training stability and poor early
decisions can lead to deadlocks and stalled learning. 
Thus, more complex settings require structured learning approaches,
including curriculum learning and hierarchical control.

Researchers and engineers can now use LineFlow to implement their specific production settings and
systematically generate simulation data. It enables both the testing of custom policies and the
training of RL agents. The behavior of trained agents can be used for analysis, but these policies
can also be deployed in the real-world. 
LineFlow enables research in diverse directions. It supports the study of non-stationary
problems, such as machine breakdowns or processing time drifts, which are common in real-world
manufacturing. Additionally, improving reward structures for curriculum learning and advancing
algorithms could significantly reduce training times and enhance performance in complex production
settings. Another promising avenue is transfer learning, which could facilitate knowledge transfer
across related tasks for more efficient and effective solutions. Beyond RL, LineFlow serves as a
data generator, enabling analysis of production line dynamics, bottleneck prediction, and
maintenance forecasting through supervised learning. 
To advance RL research in production
control, a unified training and evaluation framework is essential for tracking
and improving the state of the art. LineFlow fills this gap and drives further
research into optimizing agent performance in complex, dynamic manufacturing
environments.

\section*{Acknowledgments}
The project \emph{LineFlow} is funded by the Bavarian state ministry of research.
TW is funded by the \emph{Hightech Agenda Bavaria}. 
We thank our research partners from Robert Bosch GmbH and DMG Mori AG for helpful discussions,
particularly Martin Roth, Dominik Böhnlein, Markus Guggemoos, and Thomas Stark.
Many thanks to Matthias Burkhardt, Cindy Buhl, Kilian Führer, Andreas Fritz, Fabian
Hueber, Lea Müller, and Edgar Wolf for their helpful suggestions.

\section*{Impact Statement}
This work introduces a reinforcement learning framework for active control of production lines,
enabling optimized operations with improved efficiency. By addressing
the challenges of real-time decision-making in dynamic manufacturing environments, our method
supports sustainable industrial practices by using existing resources of manufacturing lines in an
optimal way. This research aligns with global efforts toward smarter and greener
production systems, benefiting industries, workers, and society.

\bibliography{lineflow}
\bibliographystyle{icml2025}

\newpage
\appendix
\onecolumn
\section{An Example Use Case in LineFlow}\label{appendix:example}
To demonstrate how LineFlow can be used to model and simulate production lines,
and let an agent interact with it, 
we provide a step-by-step implementation of a simple production line. 

\subsection{Implementing the Layout}\label{appendix:implementation}
In LineFlow, a production line is represented by a class that extends the \texttt{Line}
class. Inside the \texttt{Line.build} method, we define and connect the various stations,
setting their processing times and capacities. Below is an example of a small production line, which
consists of two source stations, one supplying the main component and the other providing an
additional component required for assembly. These components are fed into an assembly station, which
combines them into a single unit. The assembled product then passes through a process station. After
processing, a switch directs the product to one of two possible output paths, each leading to a sink
station that collects the finished items.

\small
\begin{lstlisting}[language=Python]
from lineflow.simulation import Line, Source, Sink, Assembly, Process, Switch

class ShowCase(Line):

    def build(self):
        source_main = Source(
            "Source1",
            processing_time=5,
            carrier_capacity=2,
            actionable_waiting_time=True,
        )
        source_comp = Source(
            "Source2",
            processing_time=5,
            part_specs=[{"assembly_condition": 100}],
            actionable_waiting_time=True,
        )
        assembly = Assembly("Assembly", processing_time=40, NOK_part_error_time=5)

        process = Process("Process", processing_time=15)
        switch = Switch("Switch", processing_time=1)
        sink_1 = Sink("Sink1", processing_time=70)
        sink_2 = Sink("Sink2", processing_time=70)

        assembly.connect_to_component_input(station=source_comp, capacity=2, transition_time=5)
        assembly.connect_to_input(source_main, capacity=3, transition_time=5)
        process.connect_to_input(assembly, capacity=2, transition_time=2)
        switch.connect_to_input(process, capacity=2, transition_time=2)
        sink_1.connect_to_input(switch, capacity=3, transition_time=2)
        sink_2.connect_to_input(switch, capacity=3, transition_time=2)
\end{lstlisting}
\normalsize

\subsection{Interaction with Agents}
In LineFlow, agents interact with the production line by receiving observations
from the environment and taking actions to optimize the system’s performance.
The state space includes features such as buffer fill levels, processing times,
and the number of workers assigned to a station. Based on these observations,
the agent can adjust parameters like switch routing decisions, worker
assignments, or waiting times at sources.
Agents can of course be trained, but also defined as a function.
The following agent dynamically adjusts the
output buffer selection at the switch based on the minimum buffer fill level.

\small
\begin{lstlisting}[language=Python]
def agent(state, env):
    fills = np.array([state[f"Buffer_Switch_to_Sink{i}"]['fill'].value for i in [1, 2]])
    return {
        "Switch": {"index_buffer_out": fills.argmin()}
        "Source1": {"waiting_time": 5},
        "Source2": {"waiting_time": 2},
    }
\end{lstlisting}
\normalsize

To observe the simulation in action, we enable visualization by setting
\texttt{realtime=True}. This ensures that the system updates in real-time, allowing for a
clear understanding of how products move through the production line. The
simulation speed can be adjusted using the \texttt{factor} parameter, which controls the
scaling of time units. For instance, setting \texttt{factor=0.1} speeds up the process by
a factor of ten, making it easier to analyze system behavior over extended
periods

\small
\begin{lstlisting}[language=Python]
line = ShowCase(realtime=True, factor=0.1)
line.run(simulation_end=150, agent=agent, visualize=True)
\end{lstlisting}
\normalsize

\subsection{Analysing the Data}
\begin{figure}[htb]
    \centering
    \includegraphics[width=\textwidth]{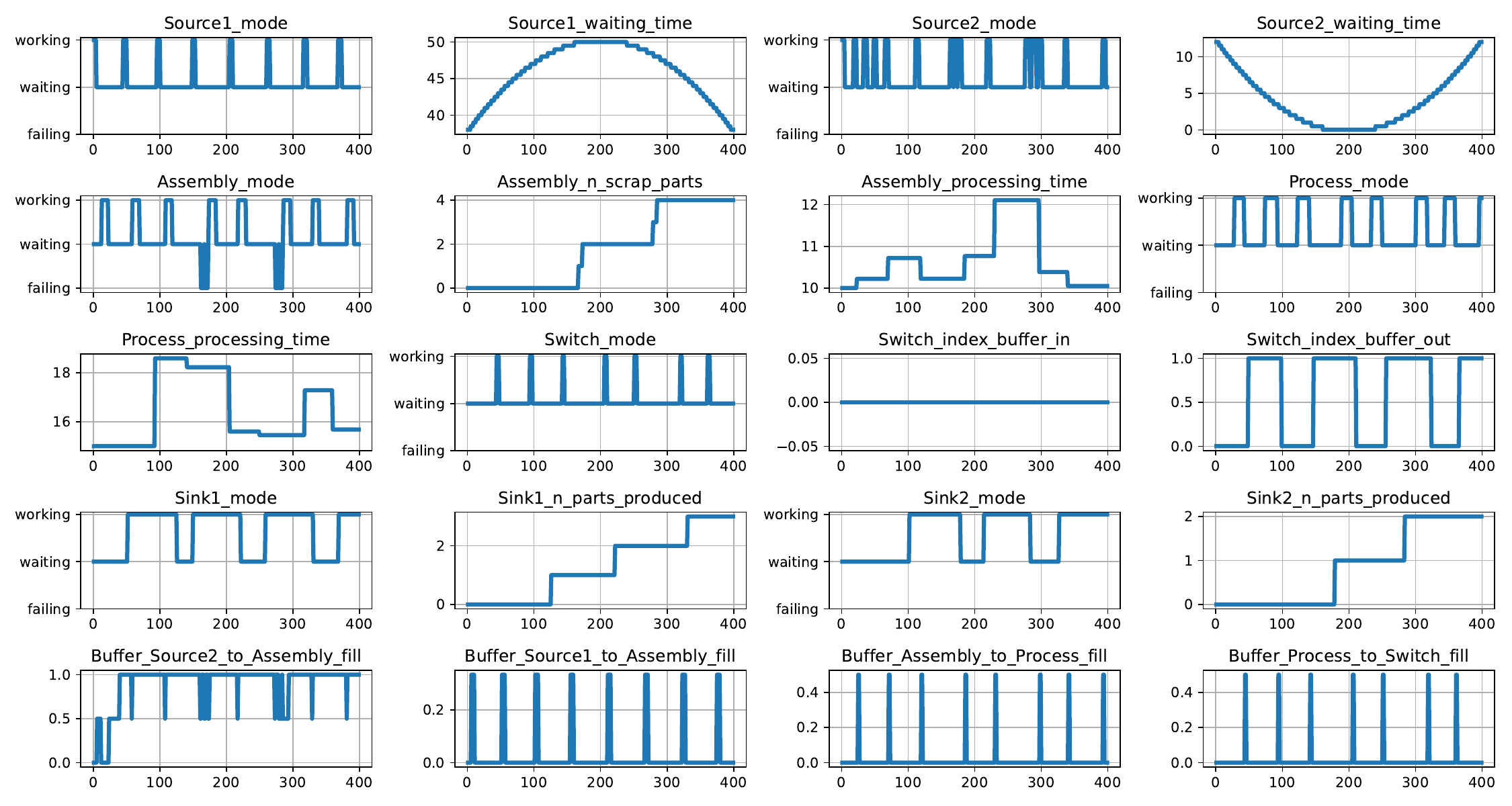}
    \caption{Selection of features extracted from the line displayed in Figure~\ref{fig:showcase}
    visualized over time.}\label{f:sample_features}
\end{figure}

The user can get all the simulation data in form of a pandas dataframe,
consisting of the states of all \texttt{lineobjects} using
\texttt{line.get\_observations()}. \texttt{LineFlow} also provides 
preinstall analysis for basic performance indicators like 
\texttt{line.get\_n\_parts\_produced()} and
\texttt{line.get\_n\_scrap\_parts()}. A sample visualization is shown in
Figure~\ref{f:sample_features}.

\subsection{Training RL Agents}
To train reinforcement learning (RL) agents, we implemented the
\texttt{LineSimulation(gym.Env)} class, which takes a \texttt{Line} object and a
simulation duration as initialization parameters. This environment integrates
seamlessly with \texttt{stable\_baselines3}~\cite{stable-baselines}, allowing RL
models to interact with the simulation by specifying it via the \texttt{env}
parameter. The following example demonstrates how to train an RL agent using
PPO:

\small
\begin{lstlisting}[language=Python]
from stable_baselines3 import PPO

line = ShowCase()
env = LineSimulation(line, simulation_time=4000)
model = PPO("MlpPolicy", env, verbose=1)
model.learn(total_timesteps=100_000)
\end{lstlisting}
\normalsize

\subsection{General Overview}\label{appendix:classinteractions}

LineFlow is implemented in an object-oriented fashion.
The central object is the \texttt{Line} object,
which is decomposed of multiple \texttt{LineObjects} (Section~\ref{appendix:stations}). A \texttt{LineObject} can either be
\emph{stationary}, like an assembly, or \emph{movable}, like a carrier. Each \texttt{LineObject} has
a \texttt{ObjectState} representing parameters of the system
a parameter. Each \texttt{ObjectState} consists of atomic states where each can be \emph{actionable}
or \emph{non-actionable} and \emph{observable} or \emph{non-observable} (see
Section~\ref{appendix:states}). The states of all objects in a line are accumulated in the \texttt{LineState} which provides a
centralized interface for \texttt{Agents} interacting with the \texttt{Line} during simulation.

\subsection{Different Types of line objects available in LineFlow.}\label{appendix:stations}

Typical properties when instantiating stationary objects are $T$ and $S$ that define their
processing time $\mathcal{T}=T+\Exp_S$ as well as a \emph{rework probability}, which models rework
by letting a process run multiple times in a row. 
Any stationary object must be connected to other objects. This is typically done via one or multiple \texttt{Buffer}
objects. For instance, a \texttt{Switch} can have arbitrarily many incoming and outgoing buffers
where an \texttt{Assembly} must have exactly one incoming and one outgoing for the main carrier and
arbitrarily many incoming for components that need to be joined with the carrier from the main
track.

\begin{longtable}{l|p{2cm}|l|p{10cm}}
\toprule
\textbf{Name} & & \textbf{Type} & \textbf{Description}\\
\midrule
\texttt{Process} &
\raisebox{-\totalheight}{\includegraphics[width=0.1\textwidth]{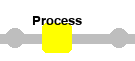}}& stationary &  A station that simulates a processing step on the part. The 
    processing step can be set to repeat due to a simulated human error, which 
    doubles the processing time. This leads to a doubled processing
time. \\
\midrule
\texttt{Sink} &
\raisebox{-\totalheight}{\includegraphics[width=0.1\textwidth]{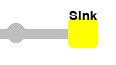}}& stationary & 
Removes components from a carrier. Carriers arrived here are marked as OK. Empty
carrier either removed or returned to a \texttt{Magazine} or \texttt{Source} using a separate
out-buffer depending on the layout. This way, the station can be used in 
    both linear and circular lines. \\
\midrule
\texttt{Source} &
\raisebox{-\totalheight}{\includegraphics[width=0.1\textwidth]{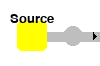}}& stationary & Places parts onto carriers. Can set individual properties, called
\texttt{PartSpec}, to every part set up, like the assembly 
condition $\Tac$. Carriers are either created, taken from a \texttt{Magazine}, or fetched separate
incoming buffer.\\
\midrule
\texttt{Assembly} &
\raisebox{-\totalheight}{\includegraphics[width=0.1\textwidth]{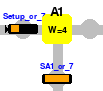}} & stationary &  A Station for simulating assembly activities on the line. 
    Individual parts and components are delivered with individual carriers, 
    assembled with a simulated processing time and forwarded to the
    downstream station. Can be connected to a \texttt{WorkerPool} that can assign  \texttt{Worker}
    objects to it modifying its processing time.\\
\midrule
\texttt{WorkerPool} &\raisebox{-\totalheight}{\includegraphics[width=0.1\textwidth]{./images/objects/assembly.png}} &  stationary & Holds a predefined set of \texttt{Worker} objects and is attached
to a fixed number of stations. Multiple pools can coexist for a production line allowing to modelling
different skills or experience of workers.\\
\midrule
\texttt{Magazine} &
\raisebox{-\totalheight}{\includegraphics[width=0.1\textwidth]{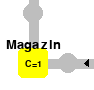}} & stationary & Magazine station is used to manage the carriers. The 
    total number of carriers available to the line can be controlled via 
    this station. The capacity of the carriers, i.e. the possible number of 
    components that can be added at the assembly station, is also determined by 
    this station. If the number of carriers is not of interest, the source can 
    place the parts directly on carriers and no magazine is required.\\
\midrule
\texttt{Switch} &\raisebox{-\totalheight}{\includegraphics[width=0.1\textwidth]{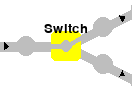}} &  stationary & The Switch distributes carriers to different stations, enabling 
    parallel structures within the line.\\
\midrule
\texttt{Buffer} &
\raisebox{-\totalheight}{\includegraphics[width=0.1\textwidth]{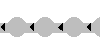}}&  stationary & The Buffer transports carriers from one station to another. Time needed to
push and get carriers to and from can be specified as well as its \emph{capacity} and time a carrier
needs to traverse the buffer.\\
\midrule
\texttt{Carrier} &
\raisebox{-\totalheight}{\includegraphics[width=0.1\textwidth]{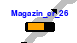}} &  movable & Is set up at a \texttt{Source} station or a \texttt{Magazine} and holds
a predefined number of \texttt{Part} objects.\\
\midrule
\texttt{Part} & \raisebox{-\totalheight}{\includegraphics[width=0.1\textwidth]{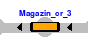}} &  movable & Single unit which is initially created at a \texttt{Source}. Holds a
\texttt{PartSpec} each station handling it can access and individually adapt to.\\
\midrule
\texttt{Worker} &
\raisebox{-\totalheight}{\includegraphics[width=0.1\textwidth]{./images/objects/assembly.png}} &  movable & Belongs to a \texttt{WorkerPool} and can be assigned to a station.
Traversal time can be configured.\\
\bottomrule
\caption{Different types of \texttt{LineObjects}.}
\label{tab:stations}
\end{longtable}

\subsection{Different Types of \texttt{States}}\label{appendix:states}

This section provides an overview over the elementary states implemented in LineFlow. 
Every state is associated to at least one \texttt{LineObject} and the mapping of selected objects to
states is given in Table~\ref{tab:mapping}.
In general, a
state can be \emph{actionable} or \emph{non-actionable} meaning that agents can set them to other
values within their value range via runtime. Per default, every state is \emph{observable}, meaning
that they are part of the observational space agents have access to. However, to facilitate ablation
studies or simulate sensor failures, any state can be selectively masked as \emph{non-observable}.
Many states are consequences of events taking place at the line, like how many carriers are on a buffer, and
cannot be influenced directly. All states kept updated constantly and represent at any time the
current situation. Some states, however, are lagged in the sense that their update happens once an
certain event is triggered. For instance, the state holding the processing time of a station is
updated once the process is over with the value of the last process.

The states can be categorized into two main categories: \emph{discrete} and \emph{numeric}. Discrete
States on the one side, handle categorical items such as station modes. Internally, the values of a
discrete state are labeled encoded to integer numbers. A subclass are \emph{count}
states that represent integer numbers, like the number of scrap parts at a station. Numeric states,
on the other side, hold continuous numerical values, like processing times or buffer fill levels.
Their value range may be restricted by upper and lower tolerances. Table~\ref{tab:states} provides a
list of states implemented in LineFlow.
All states are mapped to the respective states of gymnasium~\cite{gymnasium}.
At any simulation step $t$, the states of all \texttt{LineObjects} are fetched and put into an
agent, which returns new values for the list of actionable features to be updated accordingly.
\begin{longtable}{l|l|p{2.5cm}|p{8cm}}
\toprule
\textbf{Name} & \textbf{Type} & \textbf{Actionable}  & \textbf{Description}\\
\midrule
Mode & discrete & non-actionable &  The mode a station is currently in. Is either \texttt{working},
\texttt{failing}, or \texttt{waiting}. \\
\midrule
Processing time & numeric & non-actionable & The processing time of the last process.  \\
\midrule
Waiting time & numeric & actionable & Value used to wait till two parts produced. \\
\midrule
Station-Assignment & discrete & actionable & Discrete value denoting to which station a worker is
assigned to. Must be one of the predefined stations attached to the worker pool the worker belongs
to.\\
\midrule
\#Workers & count & non-actionable & Number of workers assigned to a station. Aggregated feature
computed from the station-assignment of all workers \\
\midrule
\#OKs and \#NOKs & count & non-actionable & Number of OK and NOK parts the station has produced so
far. \\
\midrule
Buffer-Indices & discrete & actionable & The incoming and outgoing index of a stationary object,
like a \texttt{Switch}, it should get and push components to.\\
\midrule
Buffer-Fill & numeric & non-actionable &  Relative number of carriers on a buffer.\\
\bottomrule
\caption{Examples of states.}
\label{tab:states}
\end{longtable}

\begin{table}
\begin{longtable}{l|p{1.5cm}|p{1.5cm}|p{1.5cm}|p{1.5cm}|p{1.5cm}|p{1.5cm}}
& Processing time & Current mode & Number of workers & $n_\mathrm{ok}$ & Waiting time & In and out buffer\\
\toprule
    Assembly & O, N-A & O, N-A & O, A & O, N-A & - & - \\
\midrule
    Process & O, N-A & O, N-A & O, A & O, N-A & - & - \\
\midrule
    Source & O, N-A & O, N-A & - & O, N-A & O, A  & - \\
    \midrule
    Switch & O, N-A & O, N-A & -  & - & -  & O, A\\
\end{longtable}
\caption{Default mapping of selected stationary objects to states. Letters \emph{O} and \emph{A}
    stand for observable and actionable respectively, where \emph{N} marks \emph{non-observable}
and \emph{non-actionable}.}\label{tab:mapping}
\end{table}

\subsection{Visualization}
\begin{figure}[H]
    \centering
    \includegraphics[width=0.33\textwidth]{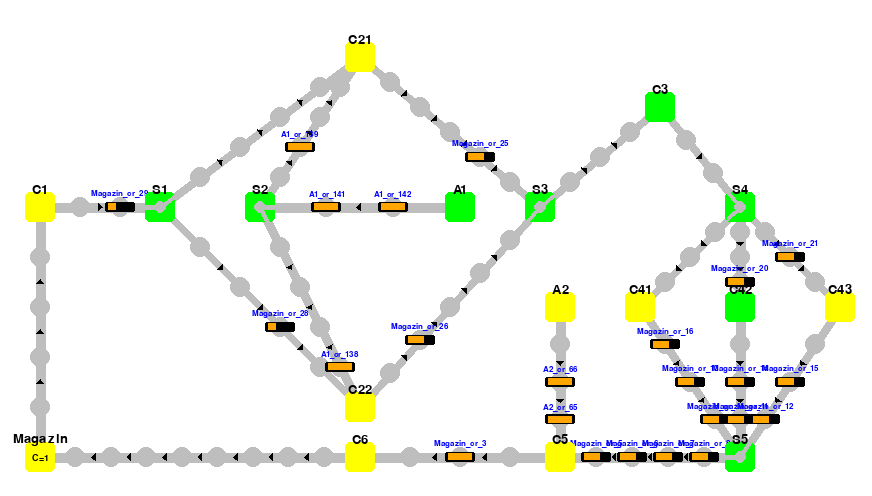}
    \includegraphics[width=0.2\textwidth]{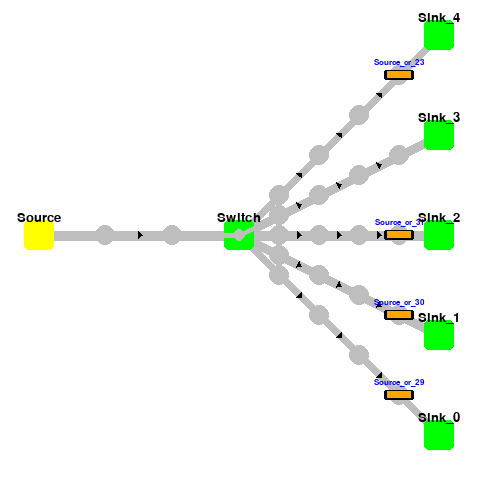}
    \includegraphics[width=0.34\textwidth]{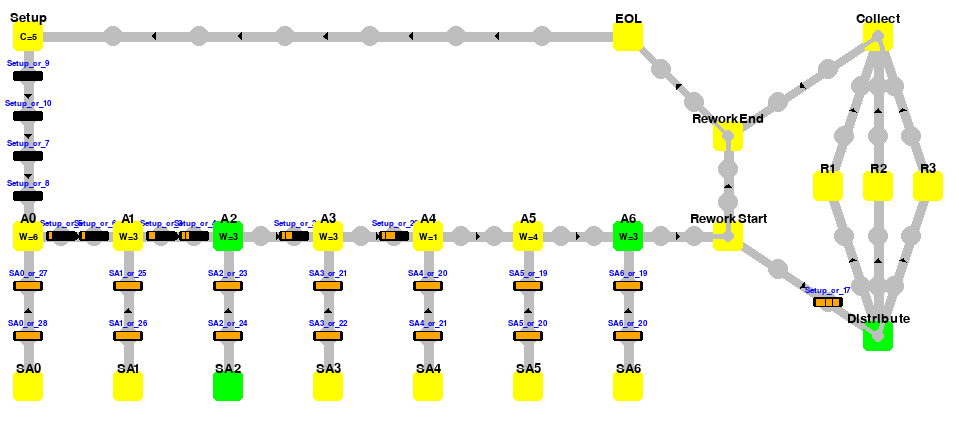}
    \caption{Some layouts implemented in LineFlow.}\label{fig:example_layouts}
\end{figure}

Once a layout is implemented in LineFlow, an object of the respective \texttt{Line} object allows to
set \texttt{visualize=True} during a simulation run, which renders the current state in pygame.
Figure~\ref{fig:example_layouts}. As layouts are implemented in custom line classes, users have
control over how these lines are instantiated and can use the power of the programming language of
python, i.e., parametrizing the number of sinks or assemblies via constructor arguments or using
for-loops to set them up. 

\section{Optimality Proofs for Case Studies}\label{appendix:optimality}

In this section, we give proofs for optimal control policies for the scenarios $\mathrm{WT}$ and $\mathrm{WTJ}$ in
Section~\ref{appendix:waiting-time}, $\mathrm{PD}$ in
Section~\ref{appendix:optimality-part-distr}, and $\mathrm{WA}$ in
Section~\ref{appendix:optimality-worker}. In Section~\ref{appendix:complex-line}, we state a
heuristic for $\mathrm{CL}$ yielding near-optimal reward.

\subsection{Waiting Time}\label{appendix:waiting-time}

\begin{figure}[H]
    \centering
    \includegraphics[width=0.7\textwidth]{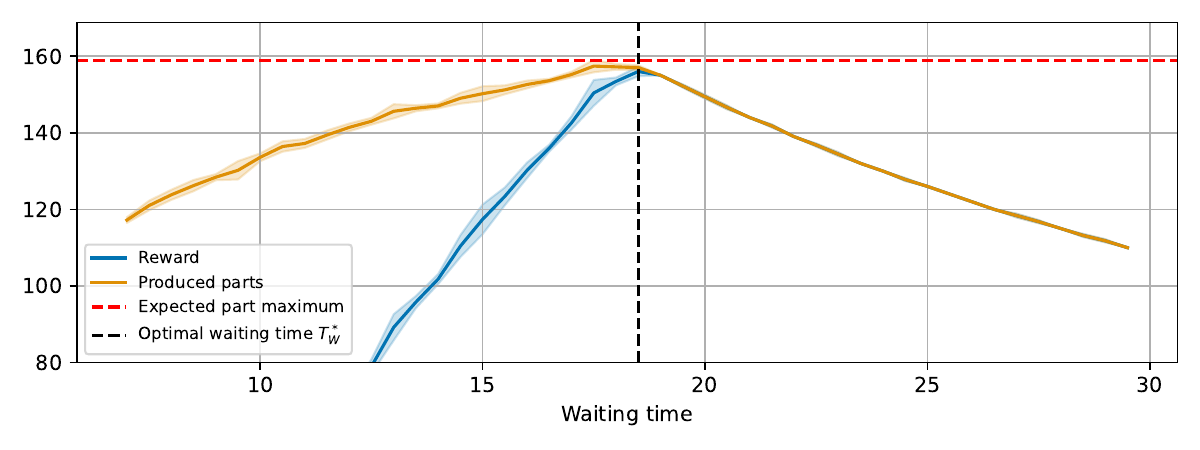}
\caption{Visualisation of the setting in $\text{WT}$ with 
optimal waiting time $T_W^*=18.5$ as in Equation~\eqref{equ:optimal-waiting-time} and the maximal number of expected
parts as computed in Equation~\eqref{equ:optimal_parts_waiting_time}.}\label{f:optimal_waiting_time}
\end{figure}

Let $\cT_{S_C}$ and $\cT_A$ be the process time distributions of $S_C$ and $A$ respectively and
denote by $\cT_p$ and $\cT_g$ the time distributions to push and get a component to and from a buffer
respectively. First, we state a formula for the optimal waiting time at $S_C$.
Essentially, the waiting time has to fill the gap between the times $A$ and
$S_C$ need to handle and process their parts, which are
$\cT_A+2\cdot \cT_g + \cT_p$ and $\cT_{S_C}+\cT_p$.
Thus, the optimal waiting time can be globally determined as the expected value of their
difference 
\begin{equation}\label{equ:optimal-waiting-time}
T_W^*=\mathbb{E}[\cT_A+2\cdot \cT_g -\cT_{S_C}].
\end{equation}
Particularly, it suffices in $\mathrm{WT}$ to learn a \emph{static} waiting time.
See also
Figure~\ref{f:optimal_waiting_time} for how the waiting time at $S_C$
affects the reward. To calculate the maximum
number of parts that can be produced, we have to consider the time that it take
for the assembly to start processing as well as the time the last part produced
in $\Tsim$ needs to be finished by the sink. 
Let $\cT_{S_C},\cT_{S_M}$, $\cT_A$, and $\cT_S$ be the
process time distributions of $S_C$, $S_M$, $A$, and the sink $S$ respectively.
Moreover, let $T_{S_C\to A}$, $T_{S_M\to A}$, and $T_{A\to S}$ be the
times a part needs to traverse on the buffers between $S_C$ and $A$, $S_M$ and 
$A$, and $A$ and the sink $S$. The assembly can start its processing once a part from $S_M$ and
$S_C$ has arrived, that is
$$\cT_{\to A}:=\max\{\cT_{S_C}+\cT_p + T_{S_C\to A},
\cT_{S_M}+\cT_p+T_{S_M\to A}\}$$
where $\cT_p$ and $\cT_g$ denote the times to push and get a component to and from a buffer
respectively. As the first part also needs to transfer to the sink $S$ and has to be
processed it, the expected value of the maximal parts
produced from $0$ to $\Tsim$ is:
\begin{equation}\label{equ:optimal_parts_waiting_time}
   \mathbb{E}\left[\frac{\Tsim-\cT_{\to A}-T_{A\to S}-\cT_g-\cT_S}{\cT_A+2\cdot\cT_g+\cT_p}\right].
\end{equation}

Next, we explain the construction of $f$ and $T_{\mathrm{jump}}$ of $\mathrm{WT}_J$ with a jumping
processing time of the assembly $A$. Following Equation~\eqref{equ:optimal_parts_waiting_time} and ignoring the ramp-up and jumps of the assembly station,  the expected number of
parts $N$ produced in $\Tsim$ is
$$
N=\frac{\Tsim}{\mathbb{E}[\cT_A+2\cT_g+\cT_p]}=\frac{\Tsim}{T+S+E}
$$
with $E=\mathbb{E}[2\cT_g+\cT_p]$. When the assembly processing with $f\cdot T+\Exp_S$ instead of
$T+\Exp_S$ for a period of $T_{\mathrm{jump}}$, the number of expected parts is:
$$
\frac{\Tsim-T_{\mathrm{jump}}}{T+S+E} + \frac{T_{\mathrm{jump}}}{fT+S+E}
$$
Clearly, if $f>1$, the expected number of produced parts must be smaller than $N$. Thus, we want to
construct $f$ from a sampled $T_\mathrm{jump}$ such that the
expected number of produced parts is $R\cdot N$ for a fixed constant $0<R<1$. Then, for a 
for a sampled $T_\mathrm{jump}$, we look for $f$ such that the expected
parts satisfies:
$$\frac{\Tsim-T_{\mathrm{jump}}}{T+S+E} + \frac{T_{\mathrm{jump}}}{fT+S+E}=R\cdot N$$
It is not hard to see that this is the case for
$$f=\frac{1}{T}\left(\frac{T_\mathrm{jump}\cdot(T+S+E)}{(R-1)\cdot\Tsim+T_\mathrm{jump}}-S-E\right).$$

\begin{figure}[H]
    \centering
    \includegraphics[width=0.7\textwidth]{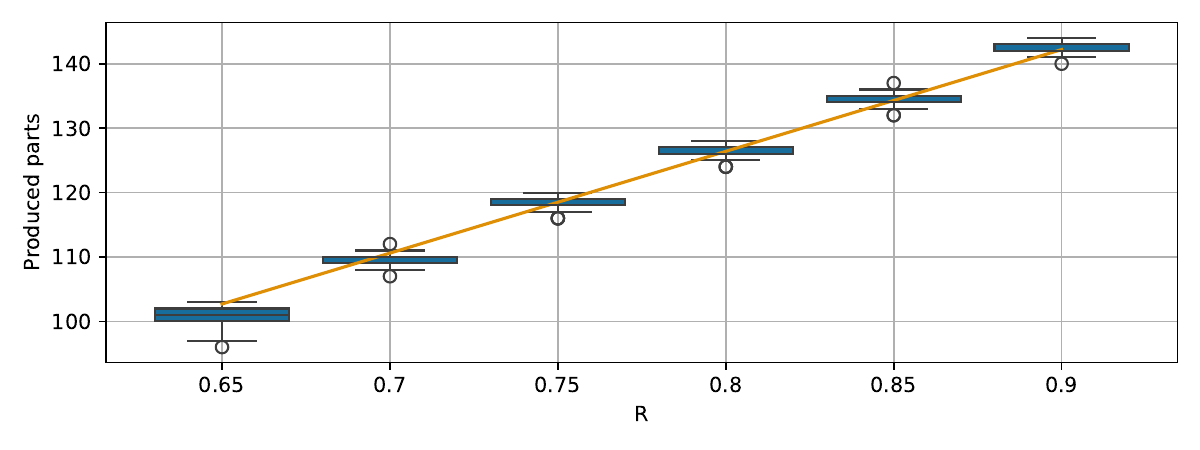}
    \caption{Number of parts produced for $\mathrm{WTJ}$ and $\mathrm{WT}$ with large assembly
    condition $T_{\mathrm{AC}}$ and no waiting at $S_C$ for varying $R$.}\label{fig:wtj_validation}
\end{figure}

To verify that the maximal number of produced parts is in fact $R\cdot N$, we run simulations in
LineFlow using varying values of $R$ and a fixed waiting time of $0$ at the source to keep the assembly
constantly loaded. As this would lead to scrap, which in turn blocks the assembly time for a certain
amount of time, we additionally set a
very large assembly condition $T_{AC}$. Figure~\ref{fig:wtj_validation} shows the empirical
frequency of the number of parts produced for $\mathrm{WTJ}$ for different $R$ values as well as
$R\cdot N$ with $N$ the expected value of $\mathrm{WT}$ showing that $f$ and $T_\mathrm{jump}$ are
constructed as desired.

\begin{figure}[H]
    \centering
    \includegraphics[width=0.7\textwidth]{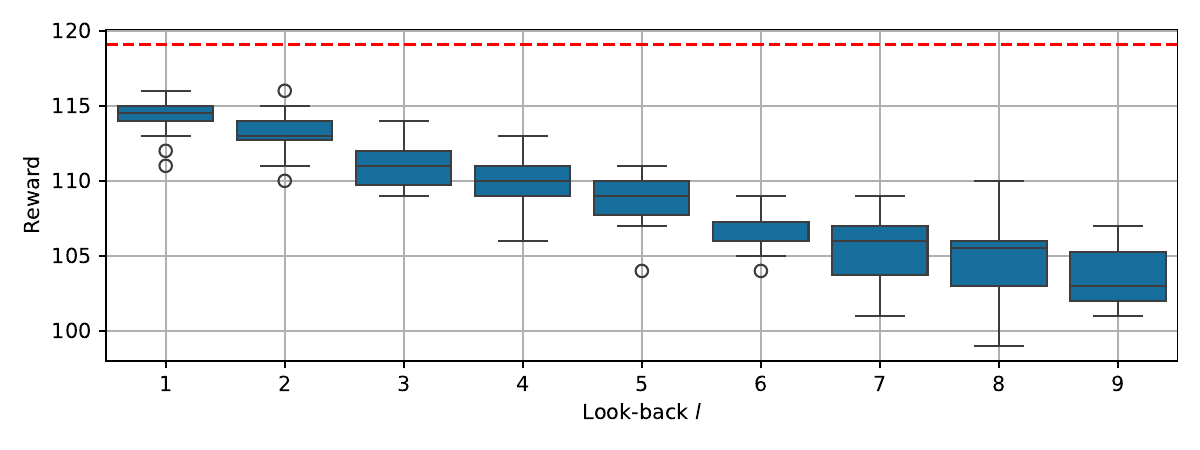}
    \caption{Reward for agent for varying look-back $l$.}\label{fig:wtj_heuristics}
\end{figure}

To get an optimal policy for $\mathrm{WTJ}$, we estimate 
$\mathbb{E}[\mathcal{T}_A+2\cdot\mathcal{T}_g-\mathcal{T}_{S_C}]$ from
Equation~\eqref{equ:optimal-waiting-time} by regressing on $\mathbb{E}[\mathcal{T}_A]$ with the
processing times reported from $A$. Essentially, we take a rolling mean of the last $l$ processing
times observations from $A$. Clearly, the larger $l$, the better the estimate for
$\mathbb{E}[\mathcal{T}_A]$ in a time period where the mean does not jump. However, the larger~$l$,
the worse the new estimate of the waiting time adjusts to a new level. By varying $l$ and testing
the agent for $\mathrm{WTJ}$ (see Figure~\ref{fig:wtj_heuristics}), we found that $l=1$ gives the
best reward. This reward is used as optimal value in Section~\ref{s:benchmarks}.

\subsection{Part Distribution}\label{appendix:optimality-part-distr}

\begin{figure}[H]
    \centering
    \includegraphics[width=0.33\textwidth]{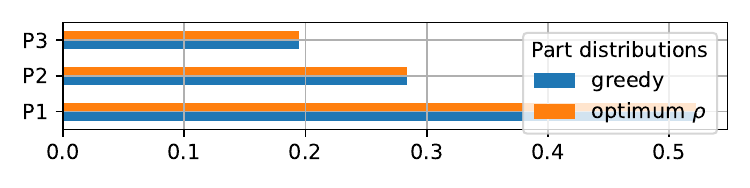}
    \includegraphics[width=0.33\textwidth]{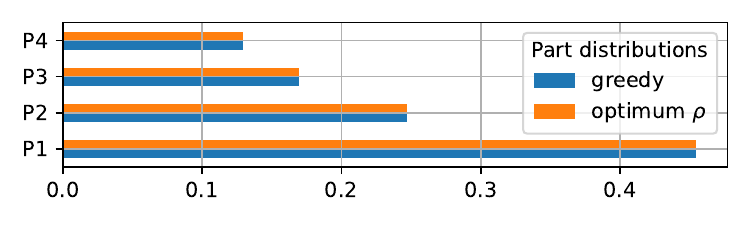}
    \includegraphics[width=0.33\textwidth]{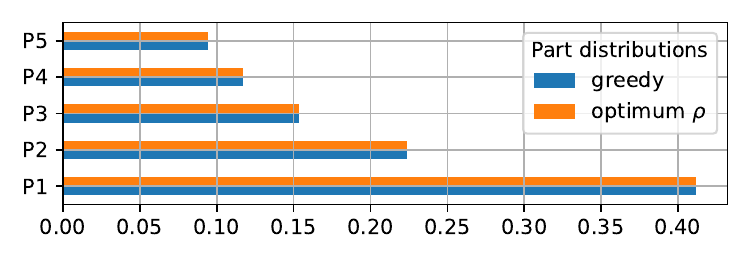}
\caption{Part distributions of the greedy policy compared with optimal distribution for
    $k\in\{3,4,5\}$.}\label{f:part-distribution-greedy-optimality}
\end{figure}

In Section~\ref{s:part-distribution}, we showed that the maximal number of parts that can be
produced in $\mathrm{PD}_k$ is
\begin{equation}\label{equ:part-distr-max-parts}
\mathbb{E}[\mathcal{N}]:=\sum_{i=1}^k\frac{\Tsim}{(1+S_i)\cdot
T_i}
\end{equation}
Particularly, the fraction of all
parts produced by $P_i$ is 
\begin{equation}\label{equ:optimal-distribution}
\rho_i=\mathbb{E}\left[\frac{\mathcal{N}_i}{\mathcal{N}}\right]=\frac{1}{\sum_{j=1}^k\frac{T_i}{T_j}}
\end{equation}
and consequently, the optimal distribution policy for the switch
also needs to deliver that number to $P_i$ to reach the maximal number
of parts. 
Put differently, the optimal policy sends a part with
probability $\rho_i$ to $P_i$.
We have implemented in LineFlow an agent that greedily controls the switches:
A carrier is pushed on a buffer having lowest fill and fetched from a buffer with highest fill.
This greedy policy have been evaluated for $k\in\{3,4,5\}$ and its part distribution is compared to
the optimal distribution $(\rho_1,\ldots,\rho_k)$ from Equation~\eqref{equ:optimal-distribution}.
The result is shown in Figure~\ref{f:part-distribution-greedy-optimality}.

\subsection{Worker Assignments}\label{appendix:optimality-worker}

\begin{figure}
    \centering
    \subfloat[$\mathrm{WA}_{3, 9}$]{
    \includegraphics[width=0.35\textwidth, angle=90]{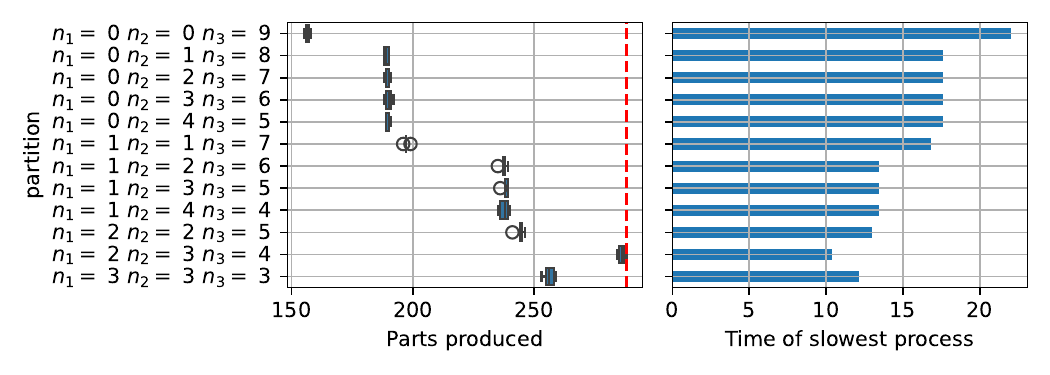}
    }
    \subfloat[$\mathrm{WA}_{4, 12}$]{
    \includegraphics[width=0.35\textwidth, angle=90]{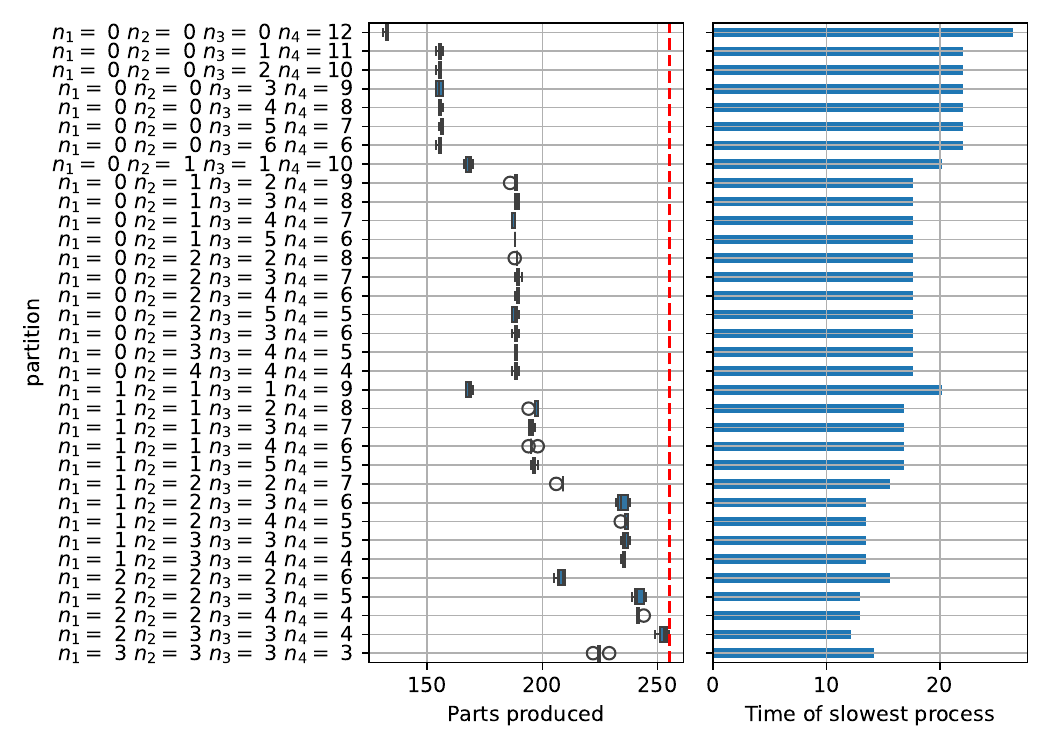}
    }
    \subfloat[$\mathrm{WA}_{5, 15}$]{
    \includegraphics[width=0.35\textwidth, angle=90]{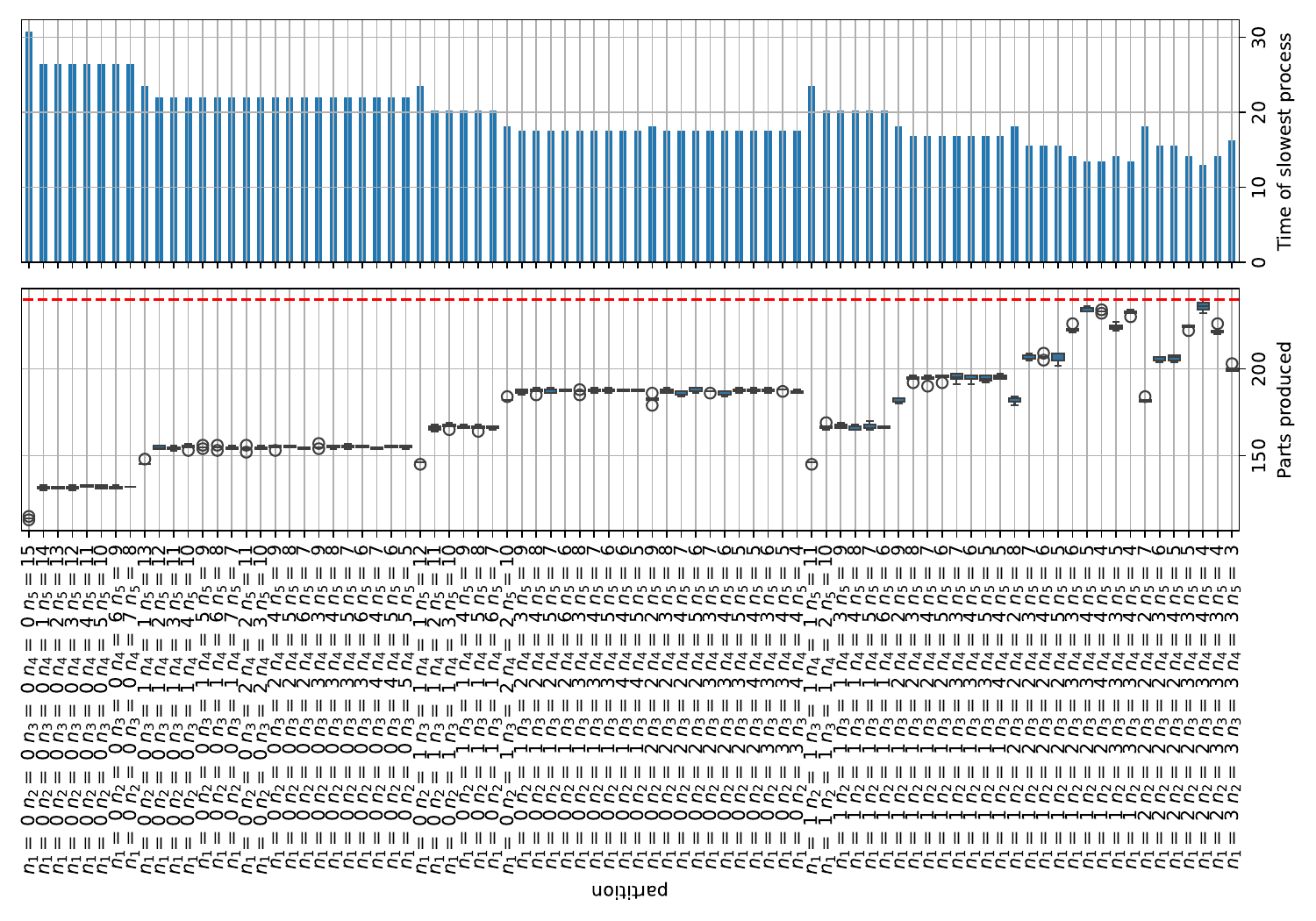}
}
\caption{Performance of all possible monotone partitions for
    $T_i=(16+i\cdot 4)$,
and $c=0.3$ for a simulation length of $2000$ in the worker distribution example.}\label{f:worker-enumerations}
\end{figure}

In this section, we formulate 
\begin{equation*}
    T_C^* = \min_{(n_1,\ldots,n_k)\in\mathcal{N}_{N,k}} \max_{i\in[k]}\mathbb{E}[\mathcal{T}_{T_i, S_i, n_i}] 
\end{equation*}
as given in Equation~\eqref{equ:worker-assignment} as integer optimization problem. 
Recall that $\mathbb{E}[\mathcal{T}_{T_i, S_i, n_i}]=T_i\cdot (p_c(n_i) + S_i)$.
We equivalently reformulate the minimization problem by introducing a real-valued
auxiliary variable $m$ as follows:

\begin{equation}\label{equ:worker-assignment-ip-formulation}
\begin{split}
&\min_{n_1,\ldots,n_k,m} m\\ 
    \text{subject to}\quad & m\ge T_i\cdot (p_c(n_i) + S_i)\quad\forall  i\in[k]\\
                           & \sum_{i=1}^k n_i = N \\
                           & n_i \in\NN_{\ge 0}\quad\forall  i\in[k]\\
                           & m\in\RR_{\ge 0}
\end{split}
\end{equation}
In our benchmarks, we used $T_i=(16+i\cdot 4)$ and $c=0.3$ for varying $k$ and $N=3\cdot k$.
We implemented this optimization problem in \texttt{gekko}~\cite{gekko}. 
Solving
Equation~\eqref{equ:worker-assignment-ip-formulation} yields $(2, 3, 4)$ for $k=3$, $(2, 3, 3, 4)$
for $k=4$, and $(2, 2, 3, 4, 4)$ for $k=5$.
To evaluate the correctness
of LineFlow, we enumerated and evaluated all monotonic worker assignment, that is $n_i\leq n_j$ for $i\le j$, for
$k\in\{3, 4, 5\}$ 
for multiple runs with simulation length
$2000$ (see Figure~\ref{f:worker-enumerations}). We found the optimal assignment
obtained empirically matches the exact optimum obtained by using
Equation~\ref{equ:worker-assignment-ip-formulation}. 

\subsection{Complex Line}\label{appendix:complex-line}
We encountered several challenges when designing an effective control policy for
the complex line. Due to the small buffer capacities between assembly stages,
the line is prone to blockages, leading to scrap parts and a significant drop in
reward. Maintaining a steady production flow proved to be crucial. 
To resolve potential jams as efficiently as possible, we implemented a switch
distribution policy that prioritizes later assembly stations. While this approach
helped mitigate blockages, it also introduced new challenges, such as imbalanced
utilization of early-stage buffers, which could lead to delays in part
availability. Regarding worker distribution, our analysis showed that keeping a fixed
distribution throughout the episode consistently outperformed any of our
redistribution strategies. We attribute this effect to the time required for
worker redistribution, which likely disrupts the system's stability.
Additionally, frequent reassignments may introduce inefficiencies due to the
time needed for workers to relocate and adapt to new tasks. We also explored adaptive worker allocation strategies that dynamically reassign
workers based on buffer fill levels. However, these approaches often led to
oscillatory behavior, where workers continuously moved between stations without
improving throughput. This suggests that the overhead of frequent redistribution
outweighs its potential benefits in our setup.

\section{Benchmarking Details}\label{appendix:hyperparameters}

In our experiments, we used stable-baselines3 (Version 2.3.2). In total, $22$ models have been trained
and evaluated for every scenario, each model using three different random seeds resulting in $66$
trained models for each scenario. The runtimes vary: Roughly $14$ hours for $\mathrm{PD}_5$ are
required to train a single model, whereas only $7$ hours for $\mathrm{WTJ}$ and $2$ hours for $\mathrm{WT}$.
To train a stacked and recurrent version of PPO for scenario $\mathrm{CL}$ NVIDIA H100 GPUs were
used. A single stacked PPO model trains roughly 20 hours, where the recurrent models train for almost $6$ days.

\begin{table}
\begin{longtable}{l|p{8cm}}
\toprule
\textbf{Name}  & \textbf{Value}\\
\midrule
Simulation time $\Tsim$ & 4,000 \\
Step size $\Tstep$ & 1 \\
Rollout steps & 1,000 \\
Discount factor $\gamma$ & 0.99 \\
Number of environments (vectorized) & 5 \\
Number of stacked observations & 1 \\
Advantage normalization (only PPO) & False \\
Batch size in update & 1,000 \\
Number of epochs in update & 5 \\
Clip range (only PPO and A2C) & 0.2 \\
Maximal gradient norm (only PPO) & 0.5 \\
Policy (stable-baselines3) & \texttt{MlpPolicy} (\texttt{MlpLstmPolicy} for recurrent PPO) \\
\bottomrule
\end{longtable}
\caption{General hyperparameters used for all algorithms.}
\end{table}

\subsection{Waiting Time}

\begin{figure}[H]
    \centering

\subfloat[$\mathrm{WT}$]{\includegraphics[height=0.15\textheight]{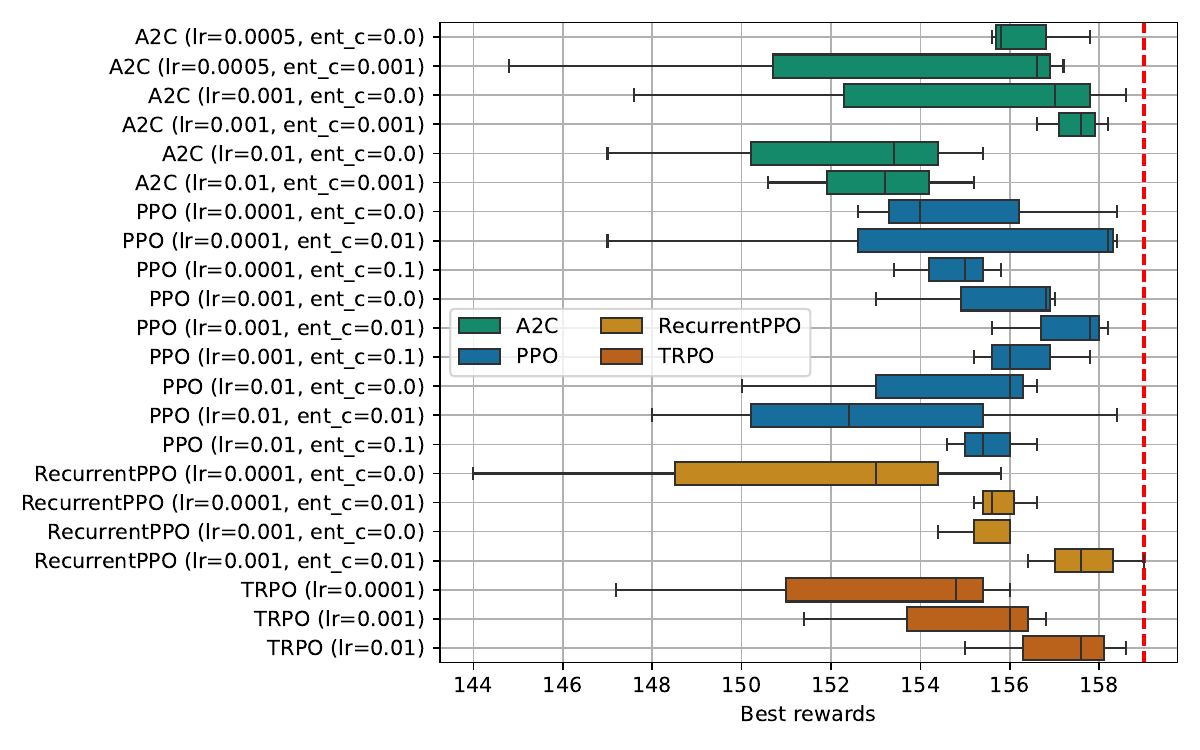}}
\subfloat[$\mathrm{WTJ}$]{\includegraphics[height=0.15\textheight]{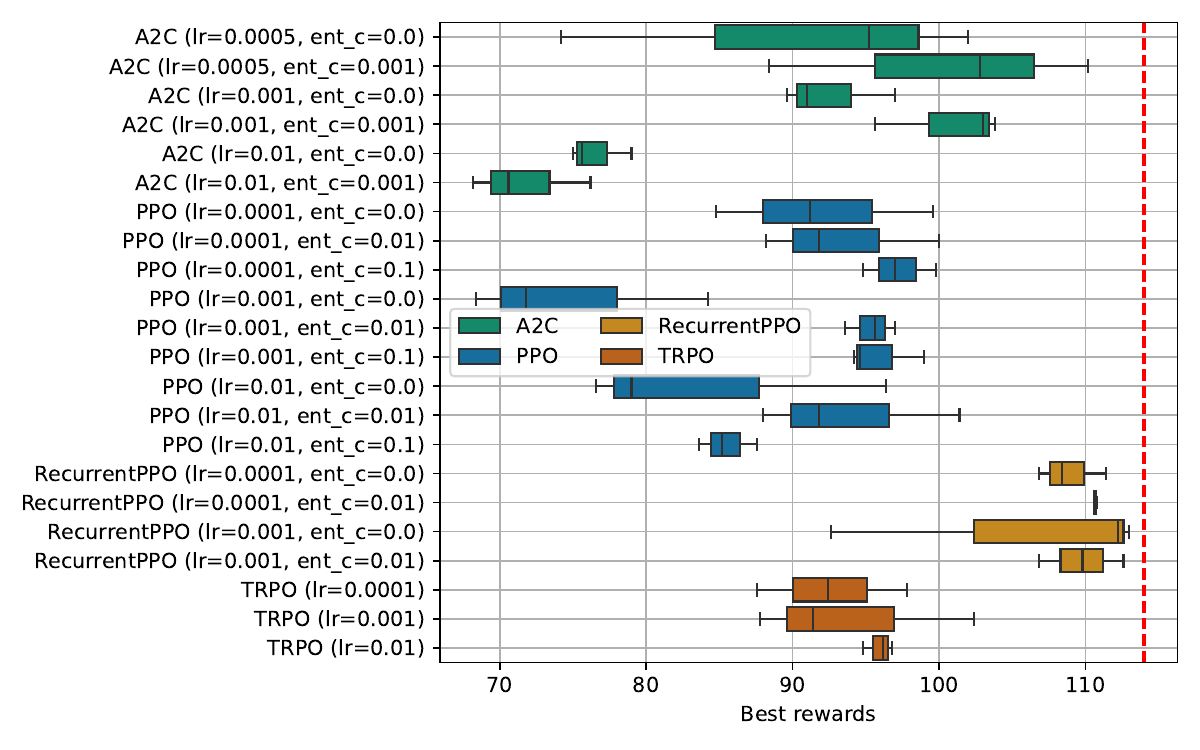}}

    \caption{Best performance of algorithms on evaluation environments for
    $\mathrm{WT}$ and $\mathrm{WTJ}$.}
    \label{f:benchmarks_waiting_time}
\end{figure}

\begin{figure}[H]
    \centering

\subfloat[$\mathrm{WT}$]{\includegraphics[width=0.45\textwidth]{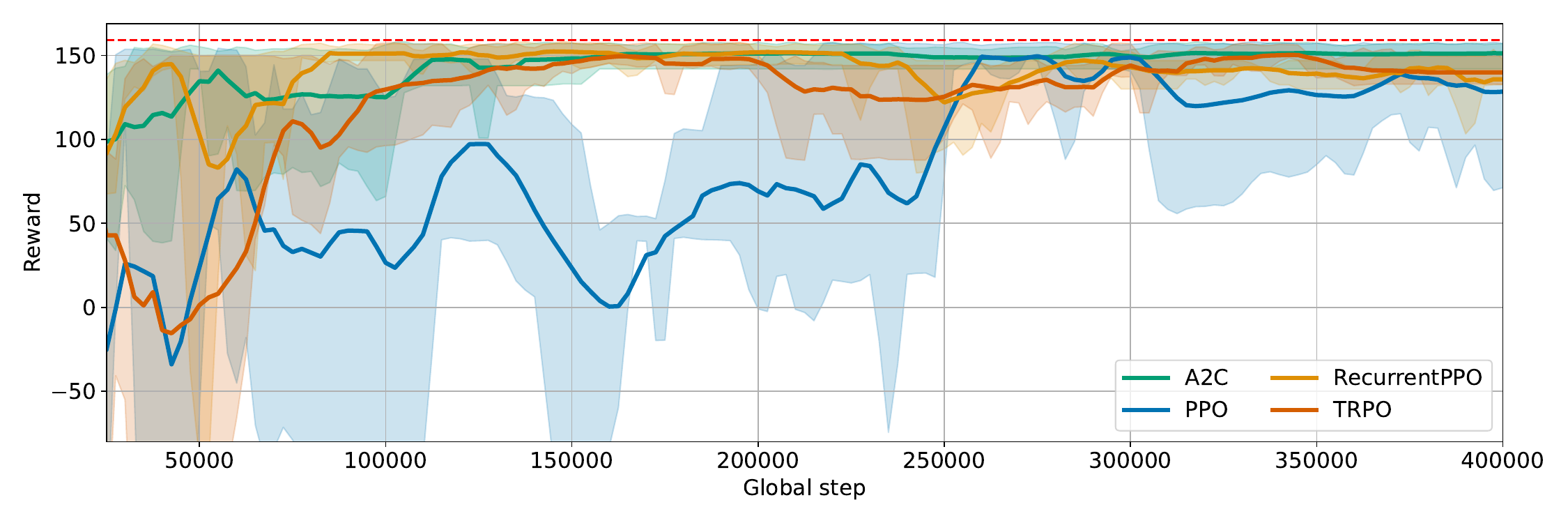}}
\subfloat[$\mathrm{WTJ}$]{\includegraphics[width=0.45\textwidth]{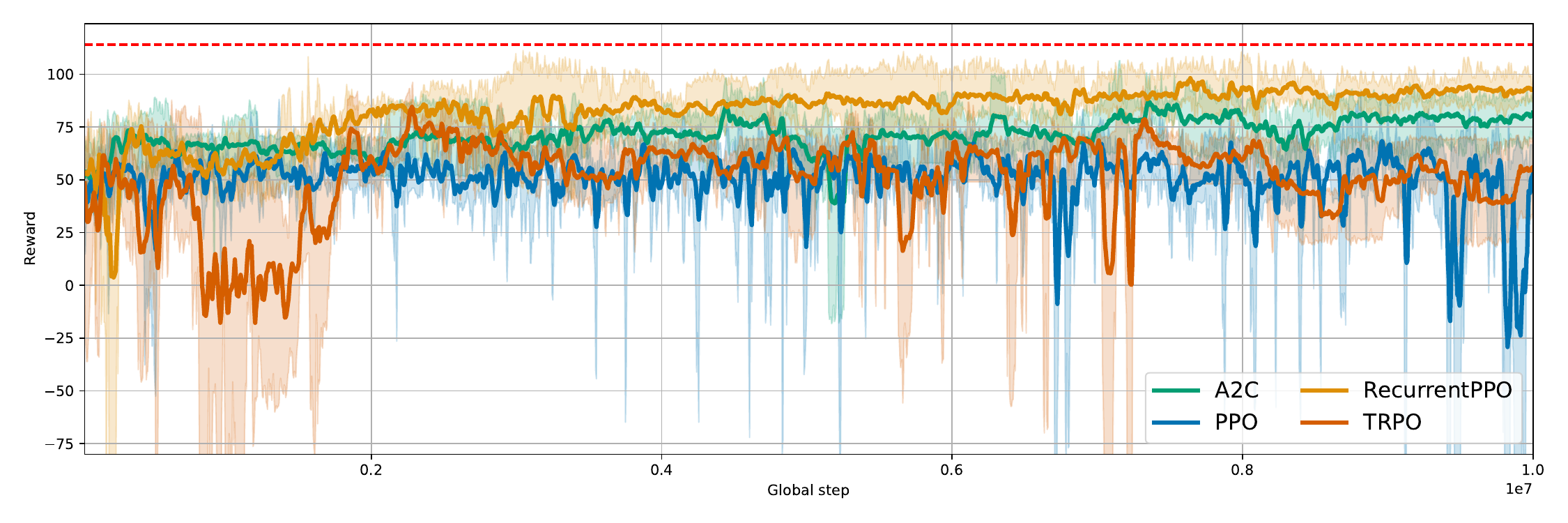}}
\caption{Reward over steps for $\mathrm{WT}$ and $\mathrm{WTJ}$.}\label{f:benchmark_waiting_time}\end{figure}

\subsection{Part Distribution}

\begin{figure}[H]
    \centering

\subfloat[$k=3$]{\includegraphics[height=0.15\textheight]{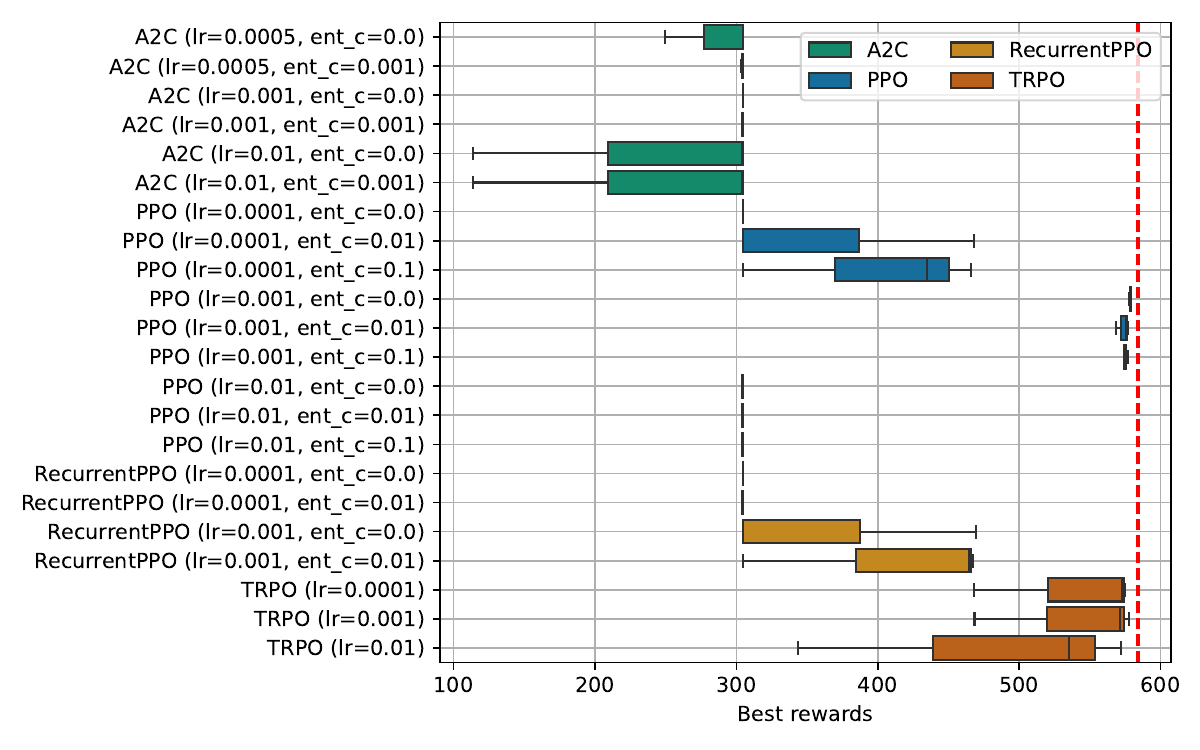}}
\subfloat[$k=4$]{\includegraphics[height=0.15\textheight]{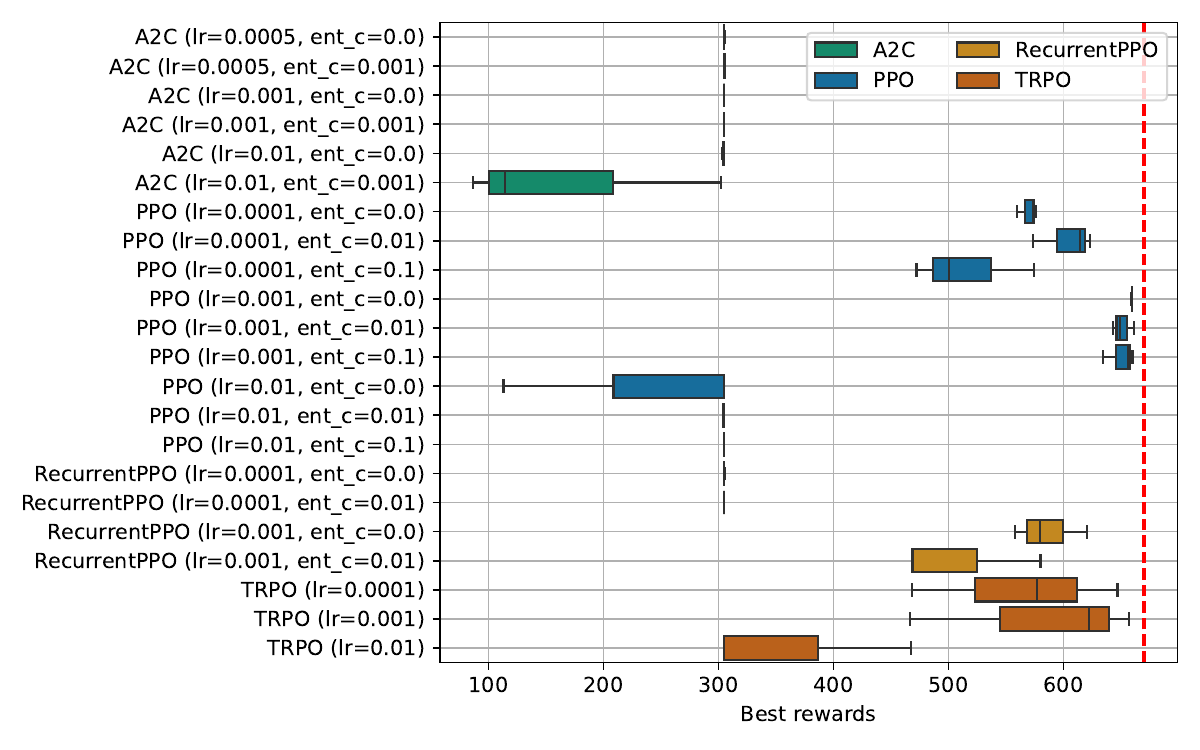}}
\subfloat[$k=5$]{\includegraphics[height=0.15\textheight]{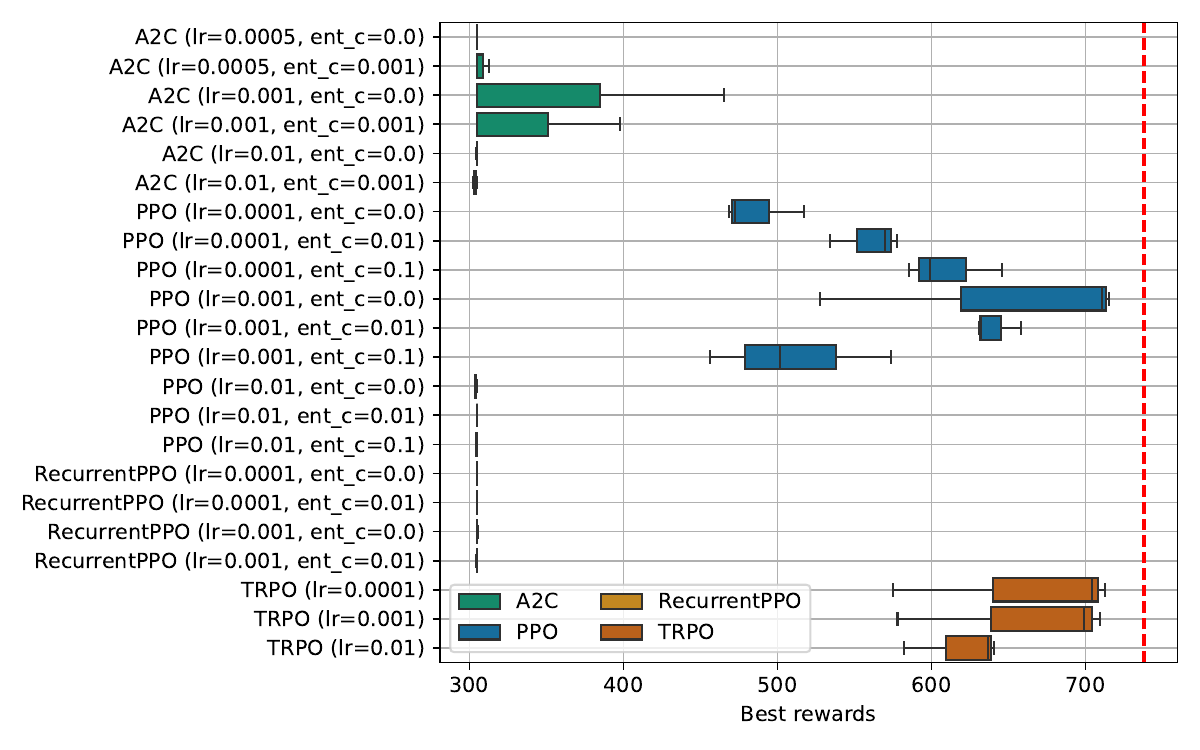}}

    \caption{Best performance of algorithms on evaluation environments for
    $\mathrm{PD}_k$.}
    \label{appendix:benchmarks_part_distribution}
\end{figure}

\begin{figure}[H]
    \centering
    \subfloat[$k=3$]{\includegraphics[width=0.30\textwidth]{./images/rl_benchmark_part_distribution_N_3_reward.pdf}}
    \subfloat[$k=4$]{\includegraphics[width=0.30\textwidth]{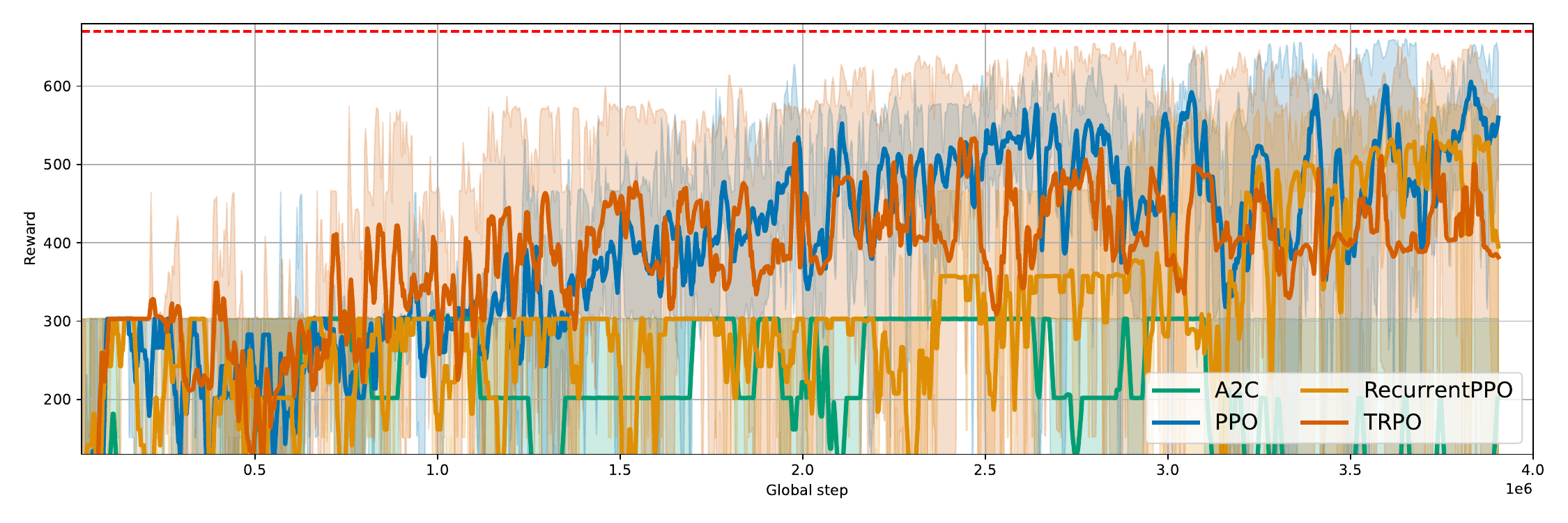}}
    \subfloat[$k=5$]{\includegraphics[width=0.30\textwidth]{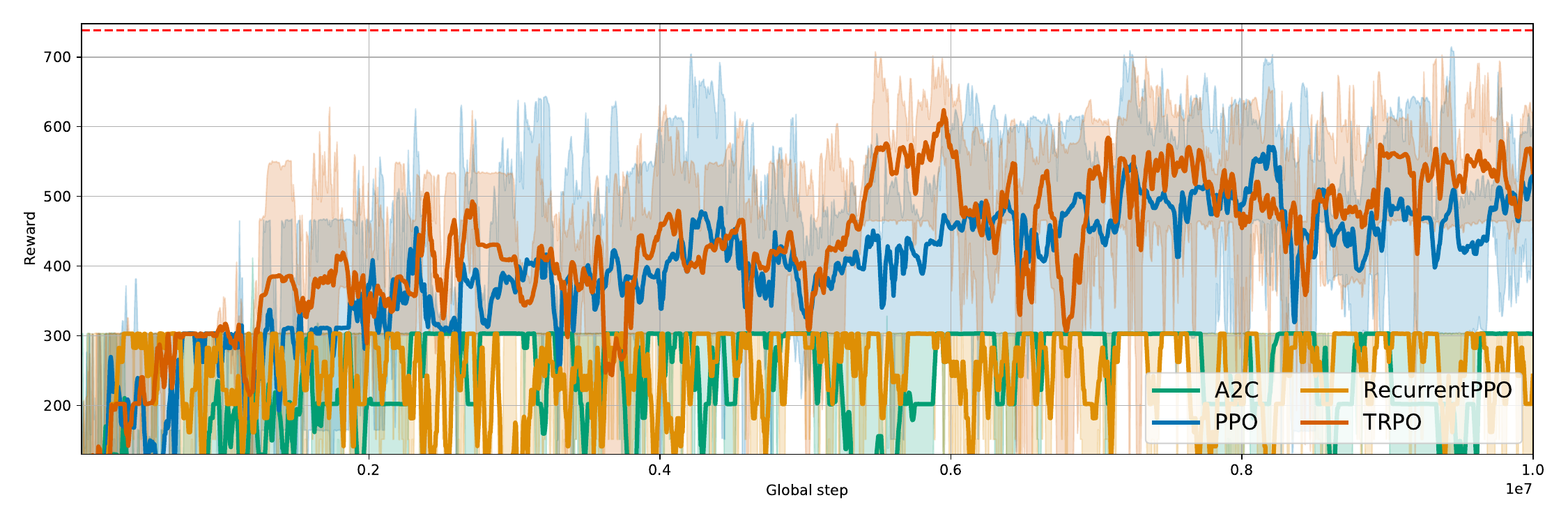}}
\caption{Reward over steps for $\mathrm{PD}_k$.}\label{f:benchmark_part_distribution_reward}\end{figure}

\subsection{Worker Assignment}

\begin{figure}[H]
    \centering
\subfloat[$k=3$]{\includegraphics[height=0.15\textheight]{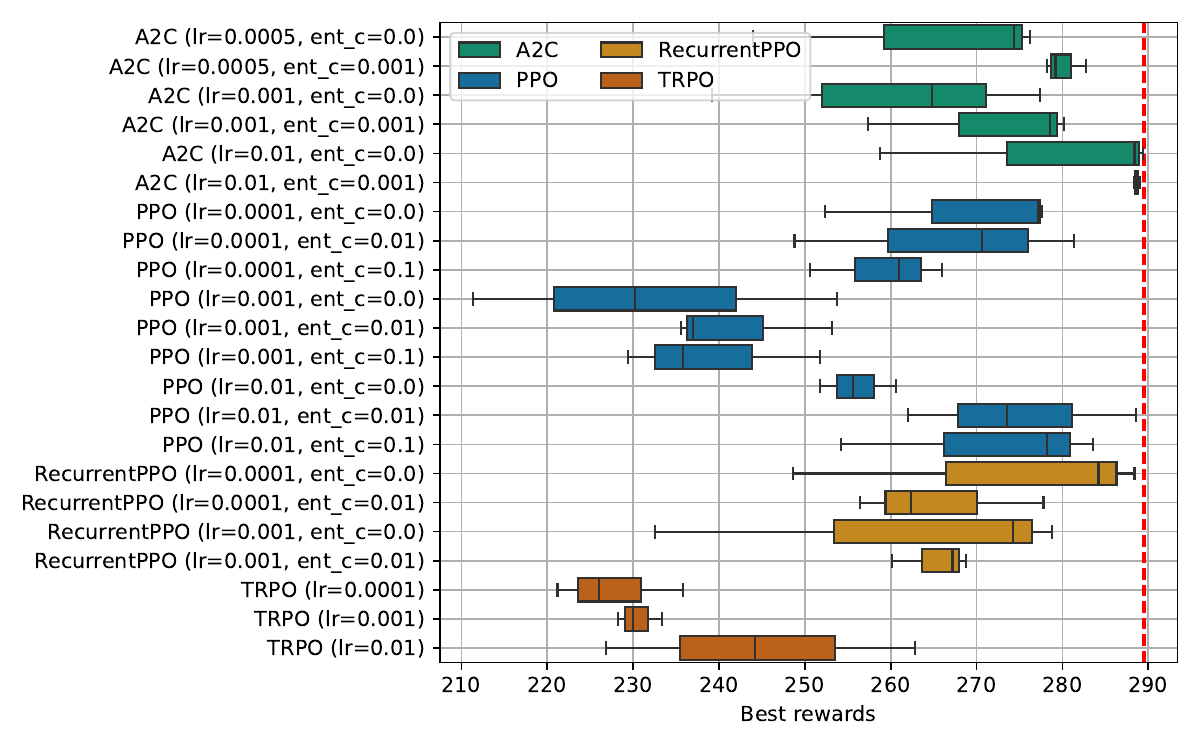}}
\subfloat[$k=4$]{\includegraphics[height=0.15\textheight]{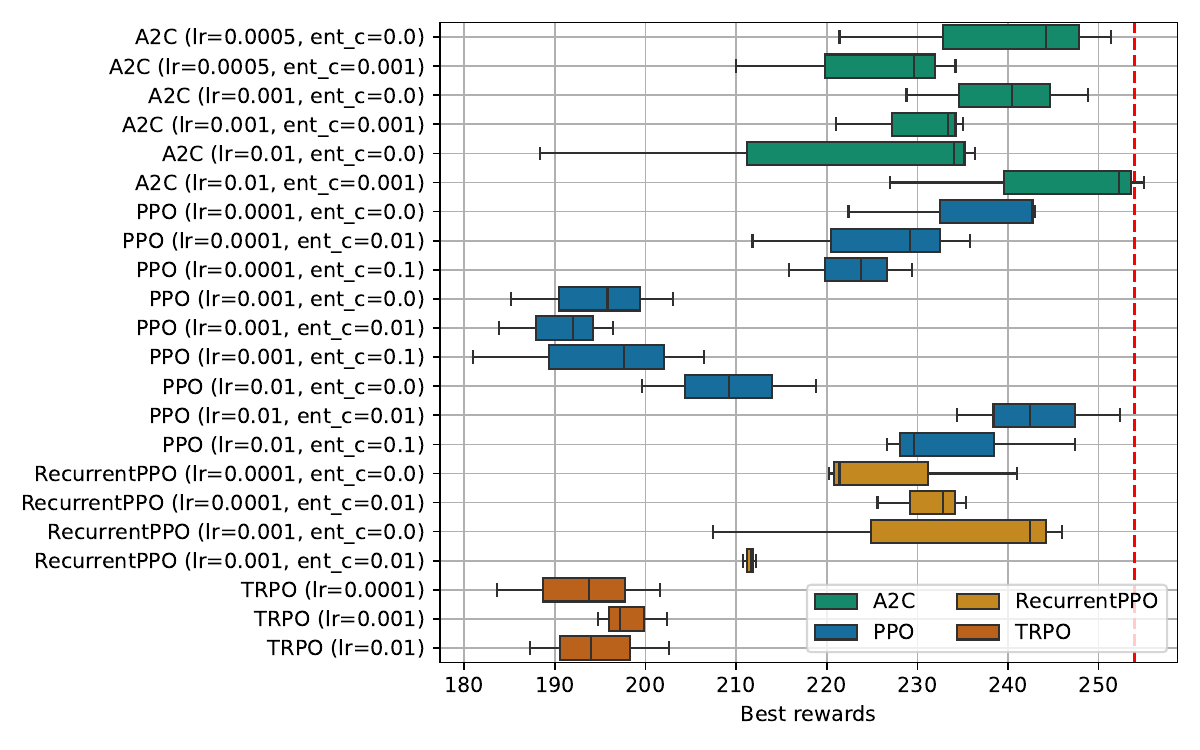}}
\subfloat[$k=5$]{\includegraphics[height=0.15\textheight]{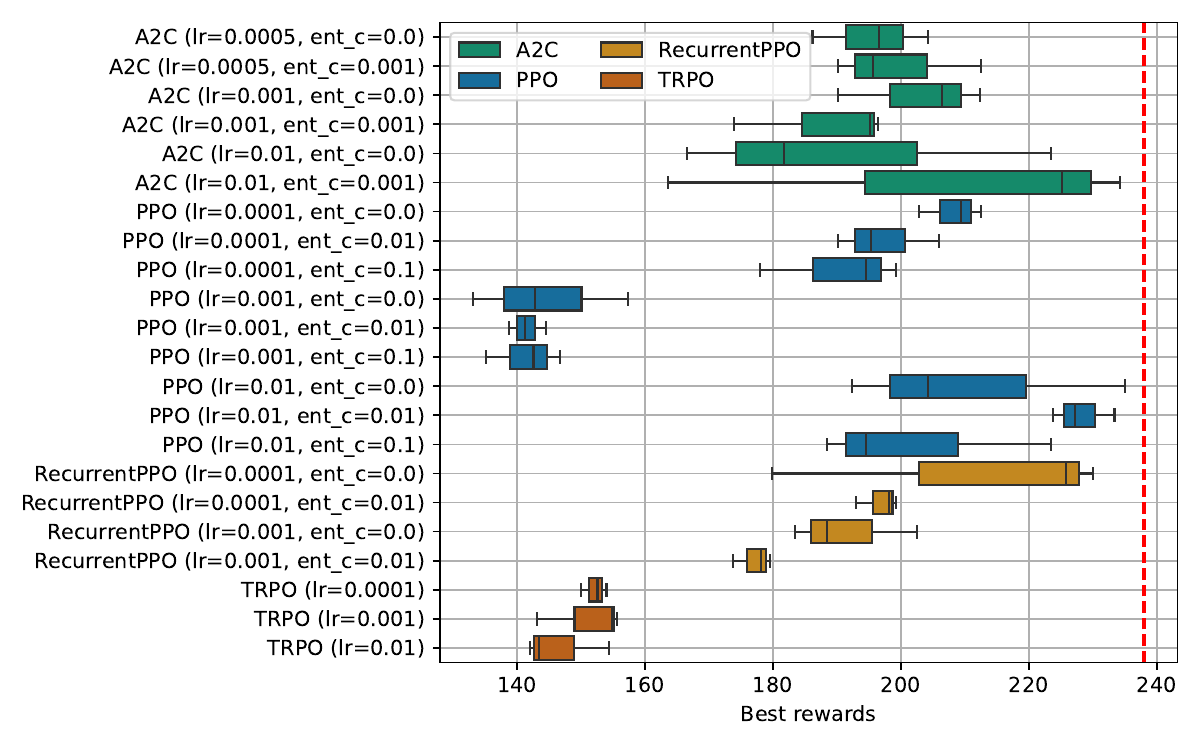}}

    \caption{Best performance of algorithms on evaluation environments for
    $\mathrm{WA}_{k,3k}$.}
    \label{f:benchmarks_worker_assignment}
\end{figure}

\begin{figure}[H]
    \centering
    \subfloat[$k=3$]{\includegraphics[width=0.30\textwidth]{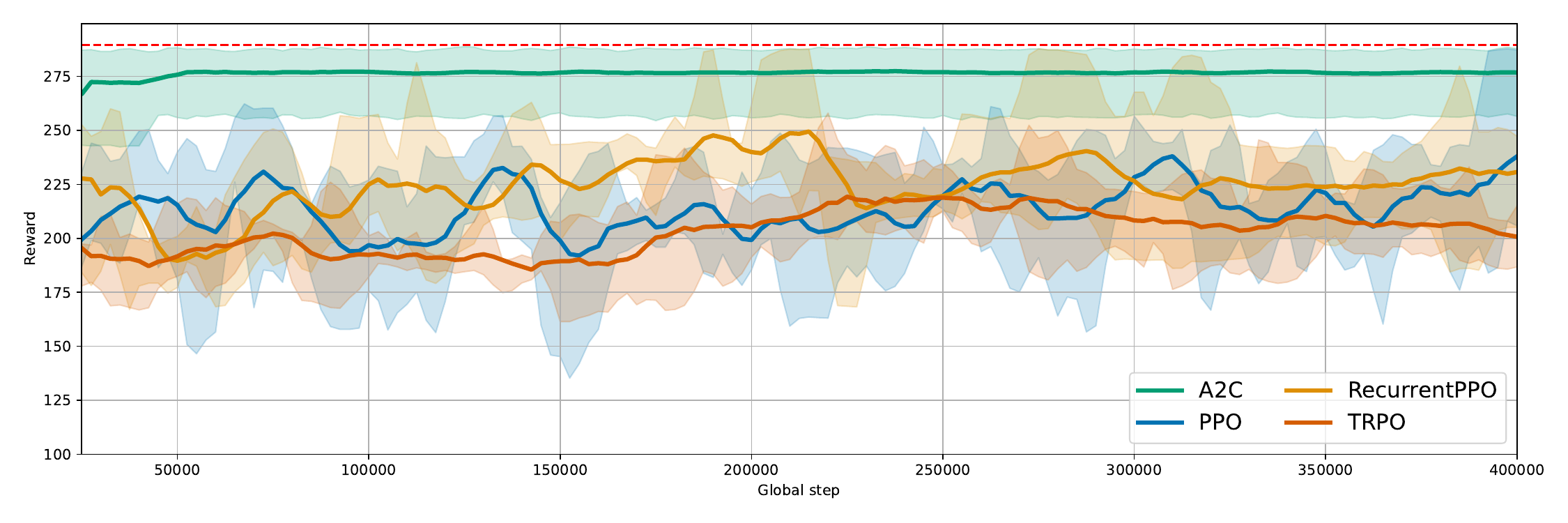}}
    \subfloat[$k=4$]{\includegraphics[width=0.30\textwidth]{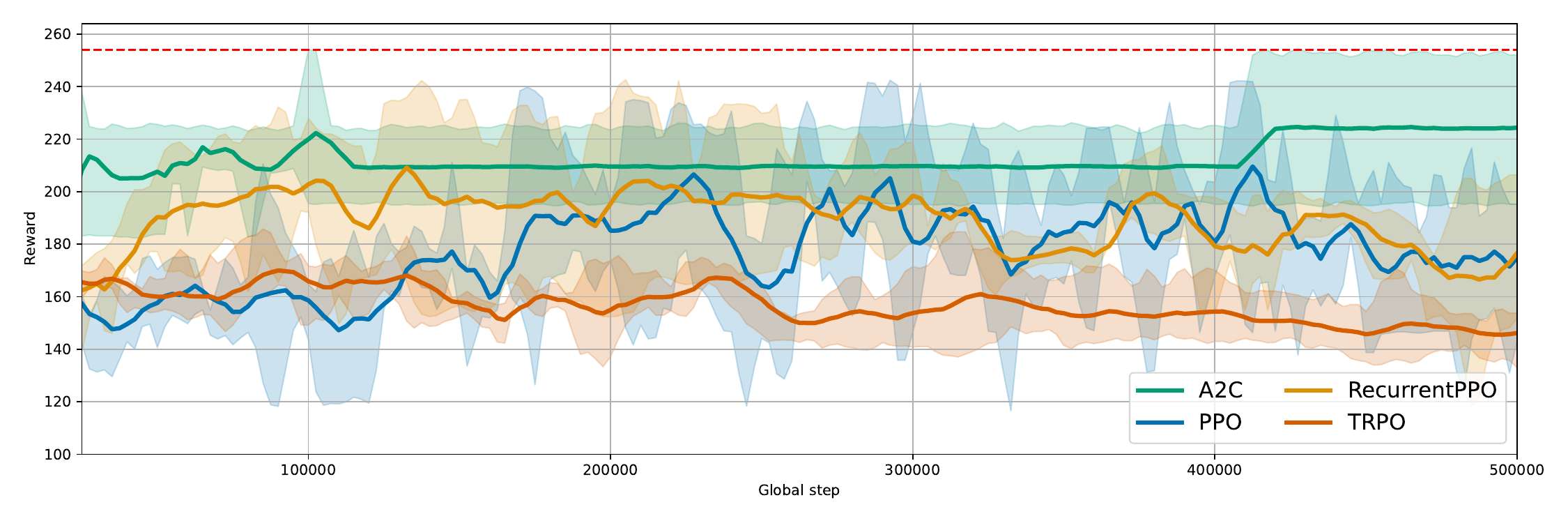}}
    \subfloat[$k=5$]{\includegraphics[width=0.30\textwidth]{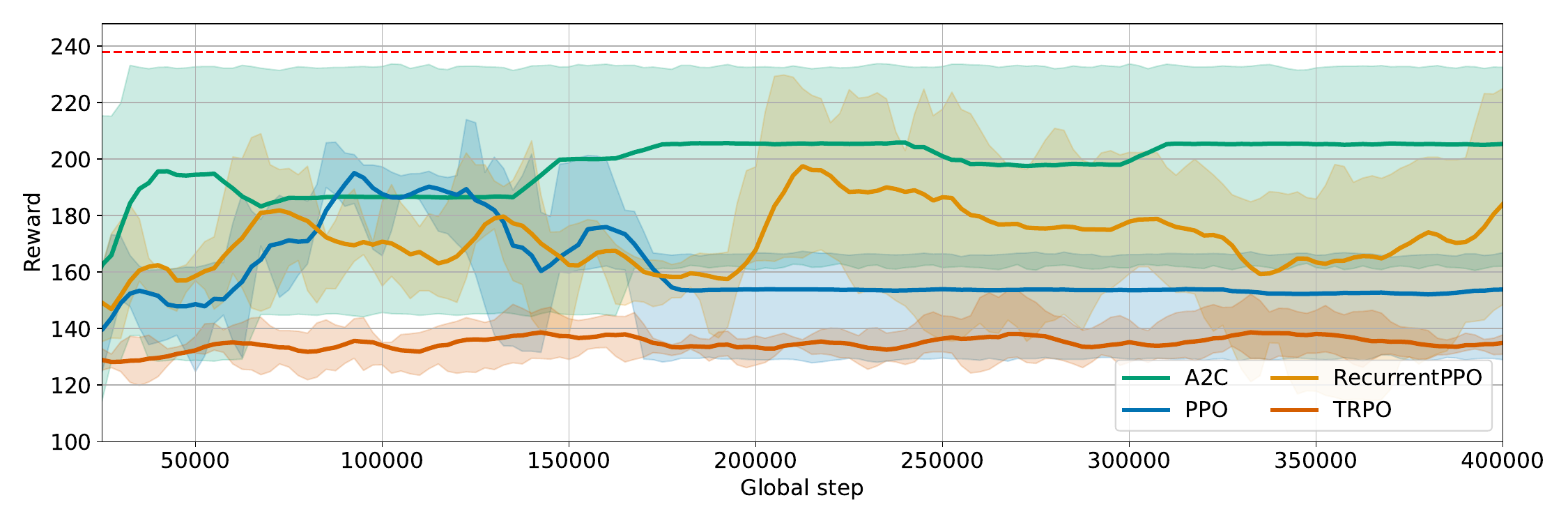}}
\caption{Reward over steps for $\mathrm{WA}_{k,3k}$.}
\label{f:benchmark_worker_reward}\end{figure}

\section{Validation on Real Production Data}\label{appendix:validation}

\begin{figure}[H]

    \centering

\subfloat{\includegraphics[height=0.20\textwidth]{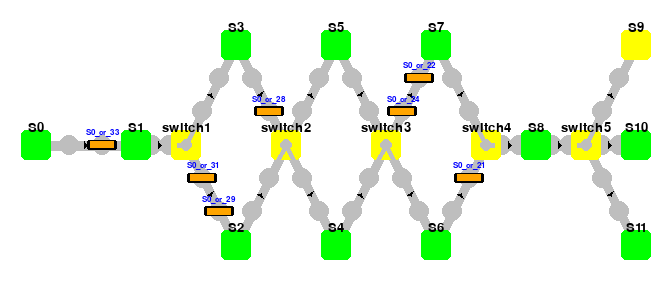}}
\subfloat{\includegraphics[height=0.20\textwidth]{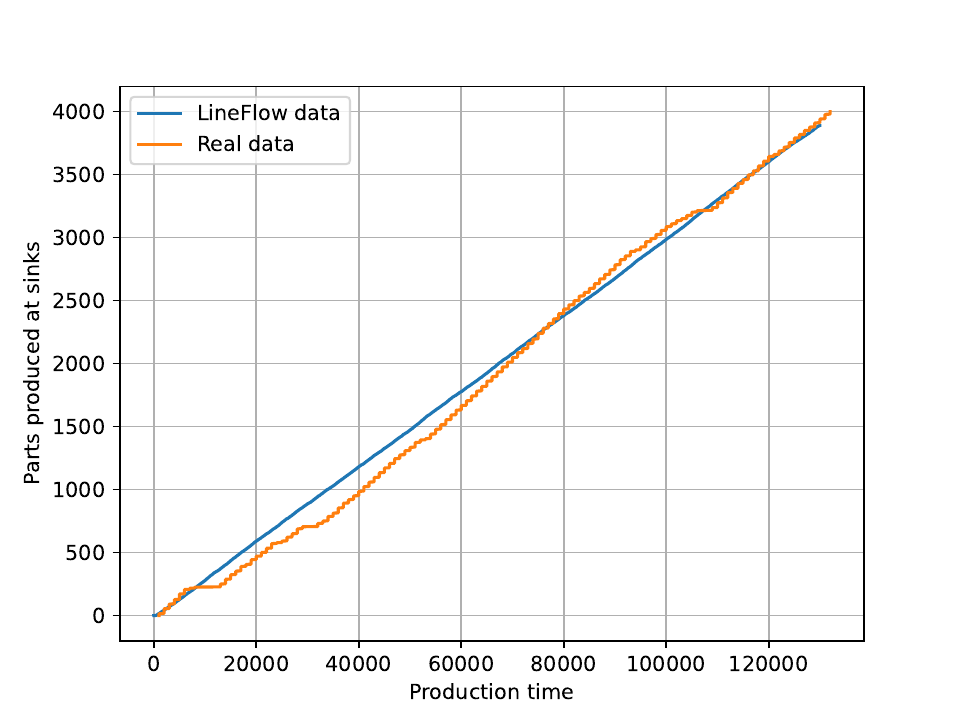}}
    \caption{The implemented layout in LineFlow (left) and its simulated output compared with the
    output of the real production line (right).}\label{fig:real_data_layout}

\end{figure}

In this section, we evaluate the sim-to-real gap of LineFlow using the publicly available
production line dataset from~\cite{bosch-production-line-performance}. In this dataset, each row
corresponds to a produced part, and the columns record timestamped feature activations at various
stations. Based on this information, we reverse-engineered the production layout, which consists of
two parallel pre-assembly lines with 12 stations each (see Figure~\ref{fig:real_data_layout}), both
feeding into a shared subsequent line. For our analysis, we focused on the pre-assembly line
responsible for the majority of component production.

\begin{figure}[H]
    \centering
    \includegraphics[width=0.6\textwidth]{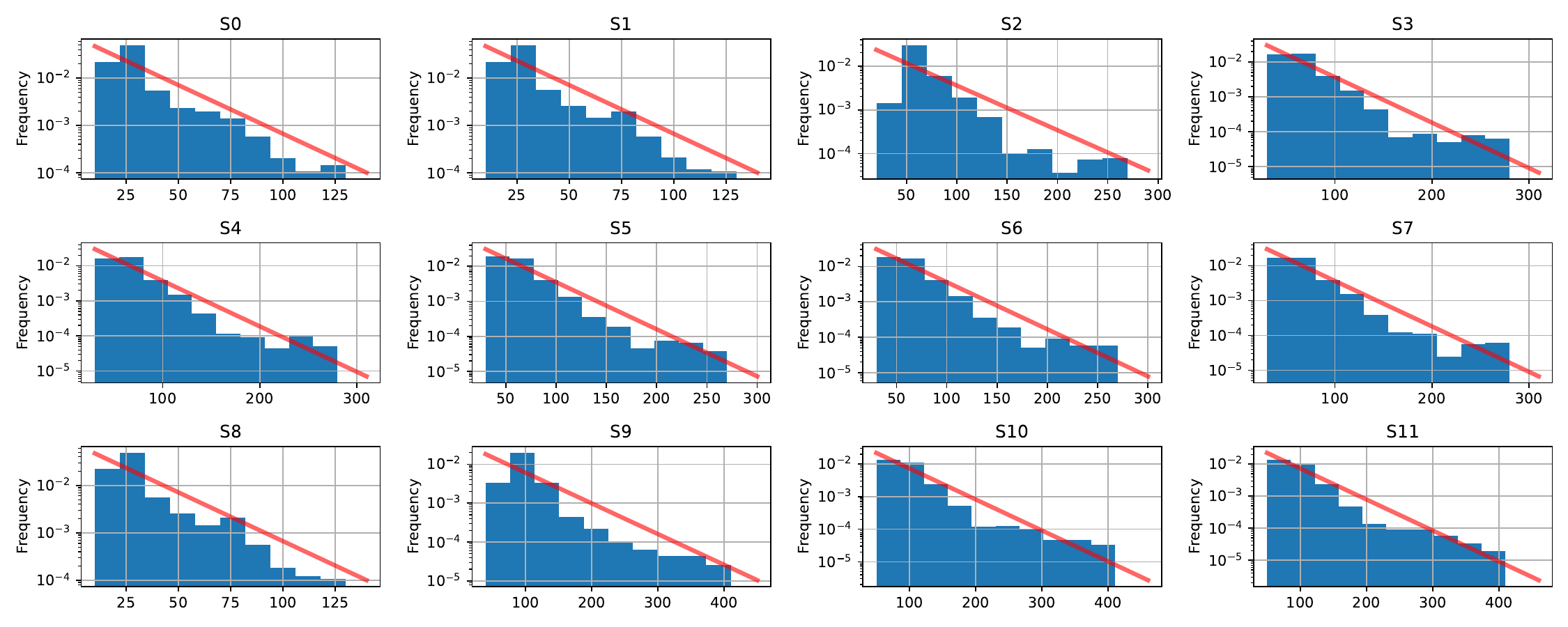}
    \caption{Histograms of the averaged processing times in a log scale together with a fitted
    exponential distribution in red.}\label{fig:exponential_distribution}
\end{figure}

We first analyzed the distribution of processing times at each station and found that they closely
follow exponential distributions, which supports the modeling assumptions used in LineFlow (see
Figure~\ref{fig:exponential_distribution}).
The limited resolution of the provided timestamps required a
pooling of the processing times of successive parts. Specifically, because individual job durations
could not be precisely resolved, we applied a rolling, non-overlapping window over the production
sequence and computed the average processing time across $100$ consecutive parts. This smoothing
technique allowed us to approximate the underlying processing time distribution while mitigating the
effects of timestamp granularity.

Next, we implemented the reconstructed layout in LineFlow and with the aim to compare the number of
parts produced in the simulation with those observed in the real system. Since the dataset does not
provide information about traversal times or the number of slots in the connecting buffer, we made
the simplifying assumption that traversal times are negligible compared to station processing times.
Additionally, we assumed that the switches before and after the stations distribute parts in a
round-robin fashion. The results showed a close match (see Figure~\ref{fig:real_data_layout}): The
number of parts produced by our simulation matches the number of parts produced by the real system,
providing empirical evidence that it accurately models real-world production dynamics.

\end{document}

%% file: images/overview.tex
\begin{tikzpicture}[scale=0.35]
    \tikzstyle{paddle}=[very thick, fill=white]
    \coordinate (O) at (0, 0, 0);

    \node[align=center] (line) at (5, 5) {
        \includegraphics[scale=0.3]{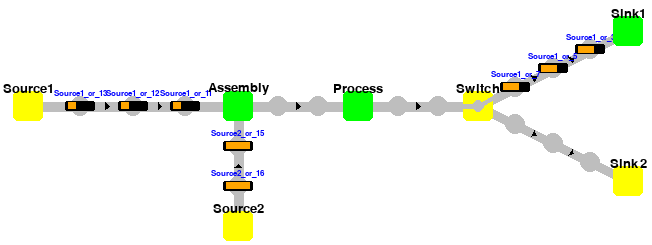}
    };

    \node[scale=0.8, align=center] (times) at (-3, -1) {Processing times\\
        \includegraphics[scale=0.15]{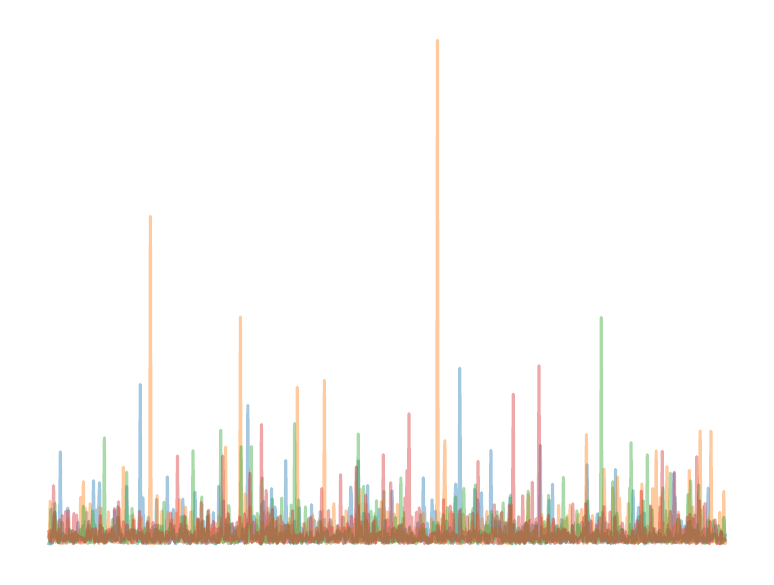}
    };

    \node[scale=0.8, align=center] (counters) at (-8, -1) {Buffer fills\\
        \includegraphics[scale=0.15]{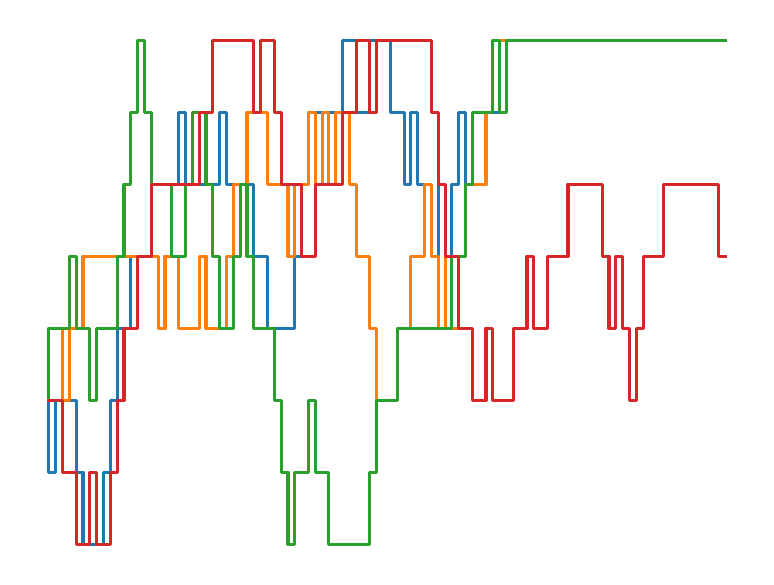}
    };

    \node[scale=2] (pi) at (7, -1) {$\pi$};

    \draw[-latex, very thick] (times) to [bend right] (pi);

    \draw[-latex, very thick] (-3,5) to [bend right] (counters);

    \draw[-latex, very thick] (pi.east) to [bend right=60] 
        node[midway, fill=white, scale=0.8, align=center] {Send parts\\to Sink2} (10,5);
\end{tikzpicture}

%% file: images/showcase.tex
\begin{tikzpicture}

\node[] at (0,0) {\includegraphics[width=0.5\textwidth]{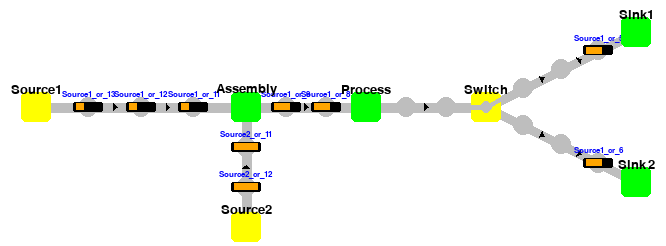}};

\node[] (buffer_1) at (1.2,0.1) {};
\node[] (buffer_2) at (2.5,0.4) {};
\node[] (carrier_1) at (-2.5,0.2) {};
\node[] (carrier_2) at (-1.2,-0.3) {};

\node[] (station_1) at (0.5,0.2) {};
\node[] (station_2) at (-1,-1.5) {};
\node[] (station_3) at (2.0,0.2) {};

\node[above=1cm of buffer_1, scale=0.7, color=blue] (buffer_txt) {Buffers};
\draw[->,thick,color=blue] (buffer_txt) to[bend left] (buffer_1);
\draw[->,thick,color=blue] (buffer_txt) to[bend left] (buffer_2);

\node[below=1cm of carrier_1, scale=0.7, color=blue] (carrier_txt) {Carriers};
\draw[->,thick,color=blue] (carrier_txt) to[bend left] (carrier_1);
\draw[->,thick,color=blue] (carrier_txt) to[bend left] (carrier_2);

\node[below=1cm of station_1, scale=0.7, color=blue] (station_txt) {Stations};
\draw[->,thick,color=blue] (station_txt) to[bend left] (station_1);
\draw[->,thick,color=blue] (station_txt) to[bend left] (station_2);
\draw[->,thick,color=blue] (station_txt) to[bend right] (station_3);

\end{tikzpicture}